%% file: main.tex
\documentclass{article}

\usepackage{fullpage}
\usepackage[T1]{fontenc}    
\usepackage[utf8]{inputenc} 
\usepackage{microtype}      
\usepackage{graphicx}
\usepackage{amsmath}
\usepackage{amssymb}
\usepackage{booktabs}       
\usepackage{multirow}
\usepackage{longtable}
\usepackage[flushleft]{threeparttable}
\usepackage{adjustbox}
\usepackage[authoryear,round]{natbib}
\usepackage[hidelinks]{hyperref}       

\title{No Universal Signal Predicts Sample-Level LLM Regression under Version Updates}

\author{
  Jia Sheng$^{1}$ \quad Yiwei Lu$^{1,2}$\thanks{Corresponding author: \texttt{yiwei.lu@uottawa.ca}.}\\
  $^{1}$University of Ottawa \quad $^{2}$Vector Institute
}

\date{}
\begin{document}
\maketitle

\begin{abstract}
Frontier LLMs are updated frequently and typically outperform their predecessors in aggregate. But aggregate gains say little about individual samples: an update can still cause sample-level regression, where a response correct under the old model becomes incorrect under the new one. This paper studies how to predict such regressions from signals available at inference time. We compare single-model signals (confidence, logit margin, attention entropy) against cross-version signals (output KL divergence, likelihood drift, token-level KL, representation drift) under a unified added-value test that isolates each signal's gain over a confidence baseline. Across six benchmarks in three task families (multiple-choice question answering, or MCQ; math reasoning; code generation) and six model update pairs, we find that (1) signal effectiveness is task-dependent: confidence is strongest on MCQ and simpler math, while likelihood/KL signals give the most frequent gains on harder math and code; (2) no signal is universally best across model updates either; and (3) some cross-version signals stay informative even when confidence fails, including without labels, which supports a proof-of-concept selective fallback that routes high-risk samples back to the old model. Practitioners can use these task-level patterns to choose which regression signal to trust for a given update. Code is available at \url{https://github.com/jiashengsally/llm-regression-signals}.
\end{abstract}

\input{sections/introduction}
\input{sections/related_work}

\input{sections/method}
\input{sections/experiments}
\input{sections/conclusion}

\section*{Acknowledgments}

We gratefully acknowledge funding support from NSERC. Resources used in preparing this research were provided, in part, by the Province of Ontario, the Government of Canada through CIFAR, and companies sponsoring the Vector Institute.

\bibliographystyle{plainnat}
\bibliography{references}

\clearpage
\appendix
\renewcommand{\thetable}{A\arabic{table}}
\renewcommand{\thefigure}{A\arabic{figure}}
\setcounter{table}{0}
\setcounter{figure}{0}
\input{sections/appendix}

\end{document}

%% file: sections/introduction.tex
\section{Introduction}

\begin{figure}[t]
\centering
\includegraphics[width=0.96\textwidth]{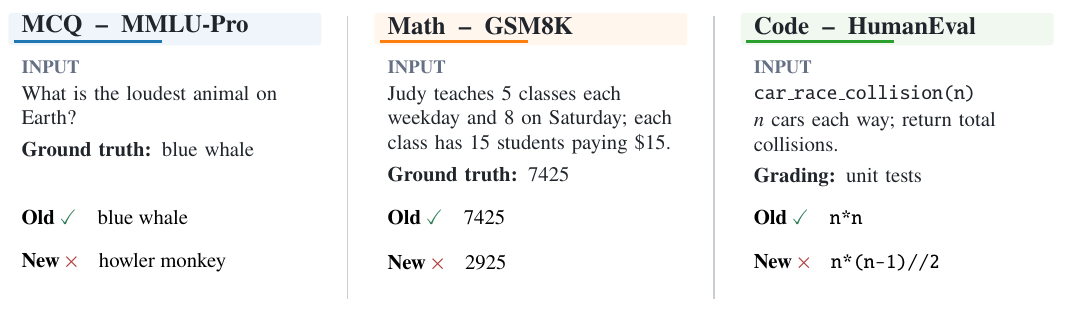}
\caption{\textbf{Representative negative flips for Qwen2.5$\rightarrow$3.} Three retrospectively selected examples, one per task family.}

\label{fig:flip-examples}
\end{figure}

Frontier LLMs are updated frequently, and each release is judged almost entirely by average accuracy (e.g., on evaluation benchmarks such as MMLU-Pro or GSM8K). But a version that scores higher overall can still fail on samples the old version got right, which we call a \emph{negative flip}. This can happen even when an update is a clear win in aggregate. Aggregate improvement and sample-level reliability therefore capture different aspects of an update. Figure~\ref{fig:flip-examples} shows representative negative flips across MCQ, math reasoning, and code generation.

In this paper, we aim to identify signals that can flag samples likely to become negative flips under version updates. A practitioner could use such signals to involve a human reviewer for high-stakes samples, or to automatically route them back to the old model as a guardrail. These signals should also be cheap to compute: rather than inspecting how model parameters changed, we study \emph{inference-time regression signals} derived only from model behavior.

We compare two families of signals for this task. Single-model signals summarize how uncertain a single model version is on a sample using only that model's own outputs: confidence (computed differently for discrete MCQ answers versus generated sequences), logit margin, and attention entropy, with the latter two specific to MCQ. Cross-version signals instead compare the old and new model on the same sample, capturing how far the update shifted the model's behavior; these include output KL divergence, likelihood drift, token-level KL, and representation drift. A natural hypothesis is that cross-version disagreement is itself predictive of regression: the more an update moves a model's answer or internal representation on a given sample, the more likely that sample is to flip from correct to incorrect. We test both signal families, and this hypothesis, under a unified added-value test that isolates their gain over a confidence baseline.

We study six model update pairs from four model families, covering about two years of real releases, with five same-family updates and one Llama-2$\rightarrow$3 boundary case (full list in Appendix Table~\ref{tab:pair-metadata}): Qwen~\citep{qwen2024qwen2,qwen2025qwen25,qwen2025qwen3}, Llama~\citep{dubey2024llama3}, Mistral~\citep{jiang2023mistral}, and Gemma~\citep{gemma2024}, all open-weight.\footnote{We study open-weight models because these signals require white-box access to hidden states, attention weights, and output logits, which closed, API-only models do not expose.} The evaluation covers six benchmarks in three task families: multiple-choice question answering (MCQ) on MMLU-Pro and GPQA, math reasoning on GSM8K and MATH-full, and code generation on HumanEval and MBPP. We do not study cross-family updates, model merging, or training-time mitigation.

Within this scope, we organize the study around four questions. (RQ1) Are negative flips simply the old model's hardest examples, or is something beyond old-model confidence needed to anticipate them? (RQ2) Which signals predict negative flips beyond single-model confidence, and does the best one depend on the task family? (RQ3) Do useful signals remain informative without ground-truth labels, and can this support a practical selective-fallback deployment? (RQ4) Can multiple cross-version signals be combined into a stronger predictor, or is one signal usually enough?

We find that signal effectiveness is largely task-dependent, and no single signal wins universally across model updates. This paper makes four contributions:
\begin{itemize}
\setlength{\itemsep}{2pt}
\item \textbf{(C1)} We show that negative flips are not fully explained by the old model's hardness or low confidence, and that this relationship weakens as tasks get harder.
\item \textbf{(C2)} We identify a task-dependent pattern in which signal family best predicts negative flips, characterizing where confidence or cross-version signals give the strongest predictions.
\item \textbf{(C3)} We show that some cross-version signals remain informative without label-based filtering, and use this to demonstrate a proof-of-concept selective fallback when confidence is unreliable.
\item \textbf{(C4)} We show that combining multiple cross-version signals rarely improves over the best individual signal, indicating competition rather than complementarity.
\end{itemize}

%% file: sections/related_work.tex
\section{Related Work}

\paragraph{Regression Measurement and Mitigation.} \citet{xie-etal-2021-regression} introduced negative flips for NLP model updates, following the vision work of \citet{Yan_2021_CVPR} on positive-congruent training. \citet{cacioli2026beyond} adapted the Reliable Change Index (RCI) to measure sample-level behavioral changes via repeated sampling, arguing this is more reliable than the single-sample negative-flip rate (NFR); but RCI is a post-hoc measure, whereas we predict regression at inference time from a single greedy decode per model. \citet{echterhoff-etal-2024-muscle} instead mitigate negative flips by training a compatibility adapter between model versions, whereas ours requires no retraining.

\paragraph{Aggregate-Level Cross-Version Signals.} Whether cross-version KL reliably predicts forgetting remains debated: \citet{shenfeld2025rlrazor} find forward KL a strong, consistent predictor, while \citet{chen2025retaining} find its value limited and inconsistent. Both evaluate aggregate forgetting under controlled fine-tuning rather than real updates. We instead study sample-level prediction under real version updates, and find that likelihood/KL effectiveness is primarily task-dependent, with weaker, preliminary evidence it also varies by update pair.

\paragraph{Sample-Level Forecasting.} \citet{jin2024forget} predict sample-level forgetting from associations between newly learned and upstream examples; \citet{jin2026demystifying} extend this with a low-rank task-by-example matrix. These are the closest prior work to ours, but both assume controlled fine-tuning on enumerable tasks and rely on statistics from repeated fine-tuning runs, so they do not directly apply to real version updates evaluated only at inference time.

\paragraph{Signal Toolbox.} The signals themselves are not our contribution: confidence follows the maximum-softmax baseline~\citep{hendrycks2017baseline}, logit margin follows margin sampling~\citep{scheffer2001active}, attention entropy adapts \citet{ostmeier2026headentropy}, and representation drift adapts hidden-state geometry~\citep{chen2024inside}; we evaluate them under a unified protocol rather than proposing new ones. This sits within a broader, unevaluated toolbox of uncertainty and representation methods, including attention-, hidden-state-, and probability-based methods~\citep{sriramanan2024llmcheck}, logit-lens representations~\citep{belrose2023tunedlens}, and semantic-entropy and probing methods~\citep{farquhar2024semantic, kossen2024seps, azaria2023saplma, phillips2026entropy}. We also draw on dataset cartography~\citep{swayamdipta2020cartography} for hardness analysis and on backward-compatible deployment~\citep{ricci2026mpt, cai2023gatedfusion} to motivate the selective-fallback strategy.

%% file: sections/method.tex
\section{Method}

\begin{figure}[t]
\centering
\includegraphics[width=\textwidth]{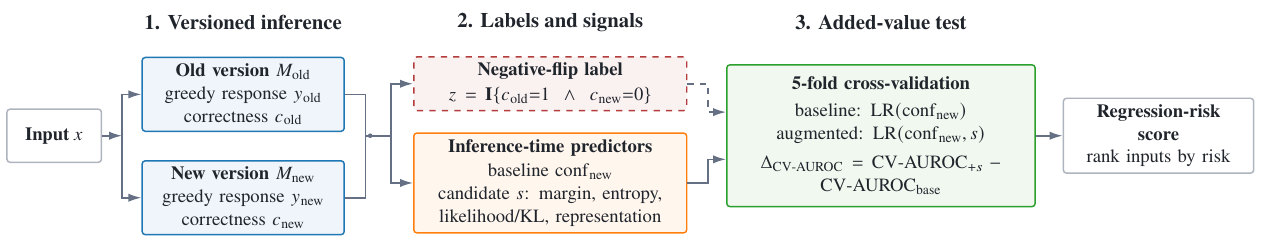}
\caption{\textbf{Method overview.} Both versions produce one greedy response per sample. The negative-flip label supervises predictor fitting and evaluation, and the signals enter a baseline-versus-augmented 5-fold cross-validation test reporting $\Delta_{\mathrm{CV\text{-}AUROC}}$.}
\label{fig:pipeline}
\end{figure}

We define negative flips and the inference-time signals, followed by the added-value test used to evaluate them. Figure~\ref{fig:pipeline} summarizes the pipeline.

\subsection{Problem Setup}

A version update replaces $M_{\text{old}}$ with a newer model $M_{\text{new}}$, usually from the same model family. Although aggregate performance may improve, some answers that were correct under $M_{\text{old}}$ become incorrect under $M_{\text{new}}$. We study these sample-level regressions.

\paragraph{Tasks of interest.}
We study three task families that differ in output structure. For multiple-choice question answering (MCQ), the answer is a single option~\citep{wang2024mmlupro, rein2023gpqa}. For math reasoning, a final answer follows a chain-of-thought (CoT) derivation~\citep{cobbe2021gsm8k, hendrycks2021math}. For code generation, correctness is decided by executing the generated program against unit tests~\citep{chen2021humaneval, austin2021mbpp}. 

\paragraph{Answer generation and evaluation.}
We generate answers using greedy decoding by setting temperature to $0$, \emph{i.e.}, a deterministic setting that selects the token with the highest probability at each step. As a result, we obtain reproducible answers and our negative-flip labels are deterministic. We use pre-specified generation budgets: $1{,}024$ tokens on GSM8K, $2{,}048$ on MATH, $512$ on HumanEval, and $256$--$320$ on MBPP depending on the update pair.\footnote{A generation that reaches its cap is truncated mid-answer and graded incorrect, which corrupts both the flip label and the trajectory-level signals; we exclude such generations from the evaluation subset.} For evaluation, MCQ answers are graded by option match. GSM8K uses the number after the final marker, with the last number as a fallback; MATH uses the last boxed answer, again falling back to the last number, followed by basic numeric and \LaTeX\ normalization. For code, we extract the generated function and run the official benchmark tests with a five-second timeout; syntax, runtime, test, and timeout failures are graded incorrect.

\paragraph{Negative flip.}
We denote per-sample correctness under the old and new models by $c_{\text{old}}(x), c_{\text{new}}(x)\in\{0,1\}$. Our prediction target is the per-sample negative-flip label
\begin{equation}
z(x)=\mathbb{I}\{c_{\text{old}}(x)=1 \land c_{\text{new}}(x)=0\}.
\end{equation}
A negative flip only occurs when the old model is correct and the new model is incorrect~\citep{xie-etal-2021-regression}. The \emph{Negative Flip Rate} (NFR) is the corresponding aggregate rate within the $M_{\text{old}}$-correct subset:

\begin{equation}
\mathrm{NFR}=\Pr[c_{\text{new}}=0\mid c_{\text{old}}=1].
\end{equation}
We assume that $c_{\text{old}}$ is given and aim to predict $c_{\text{new}}$ on the $M_{\text{old}}$-correct subset. Section~\ref{sec:label-free} later removes this filtering assumption and evaluates the negative-flip label $z$ on the full set.

\subsection{Signal Taxonomy}
\label{sec:signal-taxonomy}

Our signals differ between MCQ and generative tasks. We group them by whether they use one model version (\emph{single-model}) or compare both versions (\emph{cross-version}). Appendix Table~\ref{tab:signals} summarizes the definitions of all signals.

\paragraph{Single-Model Signals.}

We first introduce signals that quantify uncertainty using one model version (either old or new).

\vspace{0.5em}

\noindent (1) \emph{Confidence (MCQ + gen)} follows the maximum softmax probability baseline of \citet{hendrycks2017baseline}, the largest option probability computed from the option-token logits $\ell$:
\begin{equation}
\mathrm{conf}^{\text{MCQ}}=\max_i p_i,\qquad p=\mathrm{softmax}(\ell).
\end{equation}
On generative tasks it is the length-normalized sequence log-likelihood~\citep{farquhar2024semantic}, the mean per-token log-probability of the decoded sequence $y=(y_1,\dots,y_T)$:
\begin{equation}
\mathrm{conf}^{\text{gen}}(x,y)=\tfrac{1}{T}\sum_{t=1}^{T}\log p(y_t\mid x,y_{<t}).
\end{equation}
Lower confidence often indicates a higher error risk.

\vspace{0.5em}

\noindent (2) \emph{Logit margin (MCQ)} is the gap between the two largest option-token logits, frequently used in margin sampling~\citep{scheffer2001active},
\begin{equation}
\mathrm{margin}=\ell_{\text{top1}}-\ell_{\text{top2}}.
\end{equation}
A small margin means that the top two options have similar scores and indicates an uncertain prediction.

\vspace{0.5em}

\noindent (3) \emph{Attention entropy (MCQ)} adapts \citet{ostmeier2026headentropy}. For each last-layer attention head with weights $w=(w_1,\dots,w_n)$ at the final position we compute the order-2 R\'enyi (collision) entropy:
\begin{equation}
H_2(w)=-\log\sum_j w_j^2,
\end{equation}
and average the per-head entropies into a single scalar. Higher entropy means attention is spread over many positions rather than concentrated on a few, a diffuse pattern used as an uncertainty cue.  

\vspace{0.5em}

Old-model variants characterize pre-update hardness, while new-model variants characterize post-update uncertainty. We compare the relevant variants in the added-value test and analyze hardness separately in Section~\ref{subsec:hardness}.

\paragraph{Cross-Version Signals.}
Next we introduce cross-version signals that compare $M_{\text{old}}$ and $M_{\text{new}}$ on the same input and measure the change caused by the update. 

\vspace{0.5em}

\noindent (4) \emph{Output KL/JSD (MCQ)} uses the two option distributions $p_{\text{old}}$ and $p_{\text{new}}$. We compute the KL divergence~\citep{kullback1951information} in both directions and the symmetric Jensen--Shannon divergence (JSD)~\citep{lin1991divergence}:
\begin{equation}
\mathrm{KL}(p_{\text{old}}\,\|\,p_{\text{new}}),\quad \mathrm{KL}(p_{\text{new}}\,\|\,p_{\text{old}}),\quad \mathrm{JSD}(p_{\text{old}},p_{\text{new}}).
\end{equation}
A large value suggests the update substantially changed the answer distribution on the given question.

\vspace{0.5em}

The next three signals share the following scoring procedure. First, the two models greedily decode fixed output sequences $y_{\text{old}}$ and $y_{\text{new}}$. We then use \emph{teacher forcing}~\citep{williams1989learning}: both models score the same fixed sequence token by token, without generating a new response. Scoring $y_{\text{old}}$ gives the old-trajectory variant, and scoring $y_{\text{new}}$ gives the new-trajectory variant. The likelihood and token-distribution signals use the resulting probabilities, while representation drift uses the corresponding hidden states. Note that comparing token-level signals requires the same token ID to represent the same token in both model versions.\footnote{This holds for Qwen, Llama-3, and Gemma, which use identical tokenizers. Mistral v0.3 only appends 768 new tokens, so the original token IDs remain aligned; we mask the appended tokens (\emph{vocabulary masking}) when comparing it with v0.2. Updates with unrelated tokenizers, such as the out-of-scope Llama-2$\rightarrow$3, do not provide this alignment, so their token-level cross-version signals are not directly comparable.} We introduce each signal below:

\vspace{0.5em}

\noindent (5)
\emph{Likelihood drift (gen)} is the difference in per-token negative log-likelihood (NLL) between the two models on a fixed trajectory:

\begin{align}
\Delta^{\text{old}}_{\mathrm{lik}} &= \mathrm{NLL}_{\text{new}}(y_{\text{old}}) - \mathrm{NLL}_{\text{old}}(y_{\text{old}}),\\
\Delta^{\text{new}}_{\mathrm{lik}} &= \mathrm{NLL}_{\text{old}}(y_{\text{new}}) - \mathrm{NLL}_{\text{new}}(y_{\text{new}}).
\end{align}
A positive $\Delta^{\text{old}}_{\mathrm{lik}}$ means the update moved probability mass away from the old trajectory, while a positive $\Delta^{\text{new}}_{\mathrm{lik}}$ means the new model favors its trajectory more than the old model does.

\vspace{0.5em}

\noindent (6) \emph{Top-$k$ token KL (gen)} is the token-level KL between the two models' next-token distributions, averaged over the generation steps of a trajectory. At each step it is restricted to the top-$k$ tokens of the model that produced the trajectory, renormalized over that set, to avoid summing over the full vocabulary. Unlike likelihood drift, which uses only the probability of the observed token, top-$k$ token KL compares the two next-token distributions at each step before averaging across the trajectory. We use $k=50$, which retains over $0.999$ of the probability mass on average across the generative benchmarks. This choice is not sensitive within the tested range: using $k\in\{20,50,100\}$ produces highly similar signals (Pearson $r\geq0.99$) and changes $\Delta_{\mathrm{CV\text{-}AUROC}}$ by less than $0.01$.
\vspace{0.5em}

\noindent (7)
\emph{Representation drift (MCQ + gen)} adapts hidden-state geometry~\citep{chen2024inside} to the cross-version setting. It measures the cosine distance
\begin{equation}
\mathrm{rep}=1-\cos(h_{\text{old}}, h_{\text{new}})
\end{equation}
between the two models' last-layer hidden states on the same text. On generative tasks, we evaluate two variants using either the last token or mean pooling over generation tokens. On MCQ, we compare the hidden states at the final prompt position. A larger value means the internal representation moved more between versions. It requires matched \texttt{hidden\_size} and is available only for the three same-family matched pairs (Llama-3$\rightarrow$3.1, Mistral v0.2$\rightarrow$v0.3, Qwen2$\rightarrow$2.5).

\subsection{Baseline and Added-Value Test}
We select confidence (in particular, $\mathrm{conf}_{\text{new}}$ due to better empirical performance) as the baseline signal because it is simple to calculate and applies to every pair. Representation drift also applies to both MCQ and generative tasks, but it requires matched hidden sizes and is available for only three of the five update pairs. To predict $z$, we apply a simple logistic regression model to the confidence baseline, denoted as $\mathrm{LR}(\mathrm{conf}_{\text{new}})$.

To validate the effectiveness of all other signals, we perform an \emph{added-value test}. In principle, a signal $s$ is only useful when the performance of $\mathrm{LR}(\mathrm{conf}_{\text{new}},s)$ surpasses $\mathrm{LR}(\mathrm{conf}_{\text{new}})$; standalone AUROC is insufficient for this purpose, since a signal can score well simply by duplicating information already contained in $\mathrm{conf}_{\text{new}}$. Thus, we report the relative gain of every signal over this baseline rather than its standalone performance.

For evaluation, we utilize a stratified 5-fold cross-validation. The area under the receiver operating characteristic curve (AUROC) is the probability that a randomly chosen negative flip is ranked above a randomly chosen non-flip. Each logistic regression, including the direction assigned to $s$, is fit only on the training portion of its fold. We report the mean fold AUROC and define:
\begin{equation}
\Delta_{\mathrm{CV\text{-}AUROC}}
=\mathrm{CV\text{-}AUROC}_{+s}-\mathrm{CV\text{-}AUROC}_{\mathrm{base}},
\end{equation}
where a positive value indicates predictive value for the signal $s$. We say a task-level pattern is recurring when the gain is positive for a majority of the five in-scope update pairs.

%% file: sections/experiments.tex
\section{Experiments}

\begin{figure}[t]
\centering
\includegraphics[width=\textwidth]{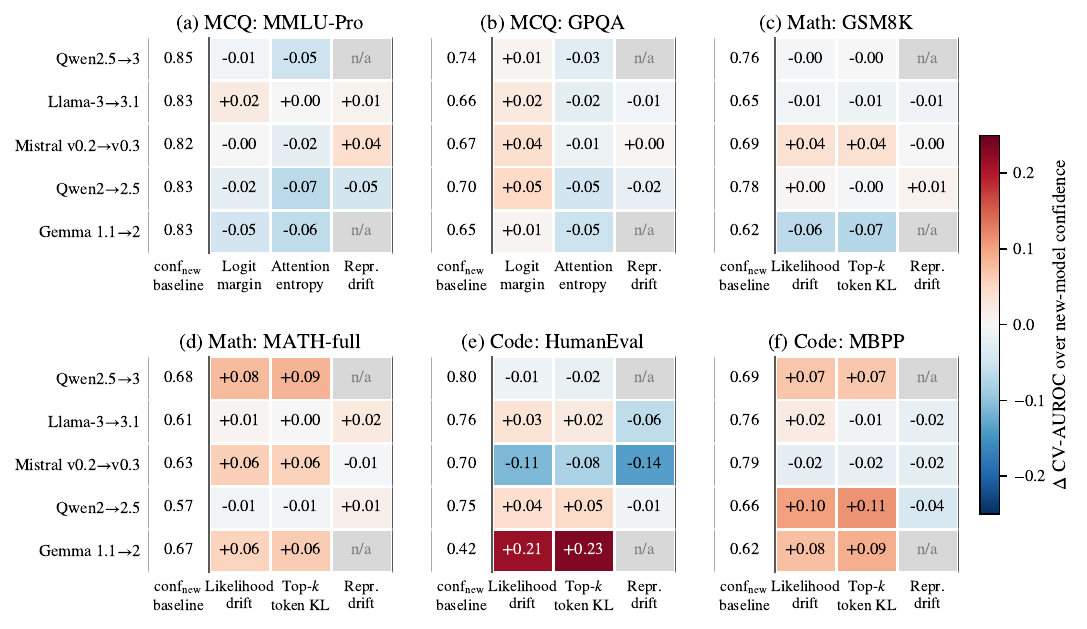}
\caption{\textbf{Signal added value across tasks.} The uncolored first column reports the absolute CV-AUROC of new-model confidence. The colored cells report $\Delta_{\mathrm{CV\text{-}AUROC}}$ after adding each signal; red indicates a gain, blue a decrease, and ``n/a'' an unavailable signal. MCQ shows old-model margin, attention entropy, and representation drift; generation shows new-trajectory likelihood drift, top-$k$ token KL, and representation drift. Appendix Tables~\ref{tab:mcq-signals}--\ref{tab:gen-code-signals} report all variants and exact values.}
\label{fig:signal-heatmaps}
\end{figure}

\subsection{Experiment Setup}

\paragraph{Models and update pairs.}
We evaluate five same-family update pairs on all three task families (see Appendix Table~\ref{tab:pair-metadata}). Every model is an instruction-tuned open-weight checkpoint with 7B--9B parameters. We do not consider models decoded with an explicit reasoning phase.\footnote{We disable thinking mode for Qwen3, the only model in our study with an explicit reasoning mode.}

\begin{figure}[t]
\centering
\includegraphics[width=1\textwidth]{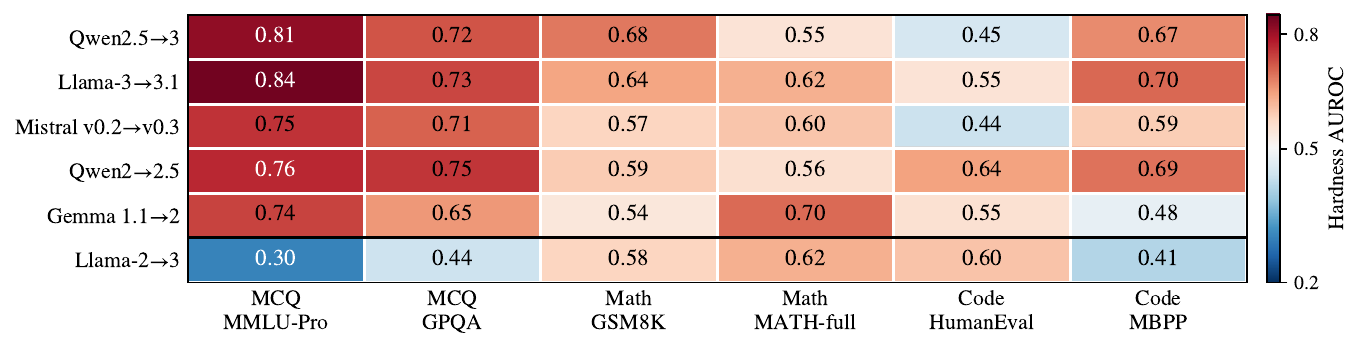}
\caption{\textbf{Are negative flips hard examples?} AUROC using low old-model confidence as flip risk on the $M_{\text{old}}$-correct subset ($0.5$ is random ranking). Red indicates that lower confidence ranks flips above kept examples; blue indicates a reversed relationship. The final row is the Llama-2$\rightarrow$3 boundary case.}

\label{fig:hardness}
\end{figure}

\paragraph{Benchmarks and validation.}
For MCQ we use MMLU-Pro~\citep{wang2024mmlupro} ($N{=}3{,}000$ per pair) and GPQA Main~\citep{rein2023gpqa} ($N{=}448$) as a secondary robustness benchmark, since its accuracy is close to the 25\% chance level and its $M_{\text{old}}$-correct subset is correspondingly small. For math reasoning we use GSM8K~\citep{cobbe2021gsm8k} ($N{=}1{,}319$) and the full MATH test set~\citep{hendrycks2021math} ($N{=}5{,}000$); our GSM8K accuracies closely match published model-card results, supporting the generation and grading pipeline. For code generation we use HumanEval~\citep{chen2021humaneval} ($N{=}164$) and MBPP~\citep{austin2021mbpp} ($N{=}500$), graded by unit tests; our single-decode protocol gives lower absolute pass rates than model-card settings, but is applied identically to both versions. Appendix Table~\ref{tab:benchmarks} summarizes the six benchmarks with their sizes, grading rules, and decoding caps.

\paragraph{Signal extraction.}
For MCQ we use direct answering and read the option-token logits from one forward pass, which keeps confidence, logit margin, and attention entropy defined consistently at the final prompt position used to predict the answer token; adding chain-of-thought would change the protocol and require separate signal definitions, so our aim is to compare signals under a common protocol rather than maximize benchmark accuracy. The generative cross-version signals use teacher forcing and vocabulary masking as defined in Section~\ref{sec:signal-taxonomy}. The final $M_{\text{old}}$-correct subsets contain $43$--$125$ HumanEval samples with $10$--$17$ negative flips per pair and $121$--$296$ MBPP samples with $17$--$40$ flips.

\paragraph{Evaluation protocol.}
The main added-value test uses the $M_{\text{old}}$-correct subset and the per-sample negative-flip label $z$. Within each cross-validation fold, the logistic regression models are fit only on that fold's training split. Each signal's direction, namely which side indicates higher flip risk, is learned only from the training split within each fold. 


\subsection{Main Added-Value Results}

Figure~\ref{fig:signal-heatmaps} reports the main results for all three task families: we include old-model margin, attention entropy, and representation drift for MCQ; new-trajectory likelihood/KL and representation drift for generative tasks. The uncolored first column reports the absolute CV-AUROC of the $\mathrm{conf}_{\text{new}}$ baseline. The colored cells report $\Delta_{\mathrm{CV\text{-}AUROC}}$ after adding each signal; red indicates a gain, blue a decrease, and ``n/a'' an unavailable signal. The Appendix tables report all variants and exact values.

\paragraph{MCQ tasks.}
Correctness on MCQ depends on one discrete choice, and $\mathrm{conf}_{\text{new}}$ measures the new model's certainty in that choice. The old-model margin and attention-entropy signals rarely add value over this baseline in Figure~\ref{fig:signal-heatmaps}. Output-distribution KL and JSD are closely tied to answer changes on MCQ, so we report them separately from the main predictive-signal comparison as direct measures of output-distribution change. \emph{In summary, new-model confidence is the strongest predictor on MCQ.}

\paragraph{Math reasoning.}
The two math benchmarks show different patterns. On GSM8K, the likelihood/KL gains are small and inconsistent, so new-model confidence is usually sufficient. On the harder MATH-full benchmark, new-trajectory likelihood drift and token KL give recurring gains for three of the five update pairs; representation drift is strongest for two matched pairs. \emph{We conclude that cross-version likelihood/KL signals help most consistently on MATH-full, and the best signal still depends on the update pair.}

\paragraph{Code generation.}
Likelihood drift and token KL give the most frequent gains on both HumanEval and MBPP, although their magnitude and direction remain update-dependent. Code correctness is functional: a fluent, high-confidence completion can still fail a unit test, and the failure may depend on any part of the generated program. The trajectory-level cross-version signals therefore add information that new-model confidence can miss. \emph{In summary, likelihood/KL signals provide the most frequent gains for code generation.}

\paragraph{Discussion.}
Overall we observe that no signal dominates across tasks. Specifically, the update pair matters, and the code sets, especially HumanEval, are small enough that pair-level gains need confirmation on larger samples. A likely explanation is task output structure: MCQ correctness depends on a single discrete choice, so new-model confidence captures most of the useful information, whereas generative correctness depends on the whole trajectory, and a fluent, high-confidence generation can still be wrong, so trajectory-level likelihood/KL signals add information beyond it. This effect is largest where confidence is a weaker proxy for correctness, namely code generation and the full MATH set, and smallest on GSM8K. \emph{We note that task output structure is a plausible explanation for this pattern, though it remains a hypothesis rather than a mechanistic one.}

\paragraph{Boundary case.}
So far we mainly focus on updates between models with comparable capabilities, which is common for successive releases in the same family. Llama-2$\rightarrow$3 (Appendix Table~\ref{tab:complementarity}) is a boundary case: the new model is much stronger and the $M_{\text{old}}$-correct subset is small, making estimates less precise.

\paragraph{Combining cross-version signals.}
We find that such combinations usually do not improve over the best individual signal (Appendix Table~\ref{tab:complementarity}). A few gains appear on the small code sets, but one well-chosen signal is usually enough, indicating competition rather than complementarity.

\begin{table}[!t]
\centering
\footnotesize
\setlength{\tabcolsep}{3.5pt}
\renewcommand{\arraystretch}{0.94}
\begin{adjustbox}{max width=\textwidth}
\begin{tabular}{@{}llccccc@{}}
\toprule
\textbf{Task} & \textbf{Signal} & \textbf{Qwen2.5$\rightarrow$3}
& \textbf{Llama-3$\rightarrow$3.1} & \textbf{Mistral v0.2$\rightarrow$v0.3}
& \textbf{Qwen2$\rightarrow$2.5} & \textbf{Gemma 1.1$\rightarrow$2} \\
\midrule
MMLU-Pro & $\mathrm{conf}_{\text{new}}$ & $0.85\!\to\!0.63$ & $0.83\!\to\!0.60$ & $0.82\!\to\!0.59$ & $0.83\!\to\!0.61$ & $0.83\!\to\!0.62$ \\
GPQA & $\mathrm{conf}_{\text{new}}$ & $0.75\!\to\!0.60$ & $0.66\!\to\!0.60$ & $0.68\!\to\!0.59$ & $0.69\!\to\!0.58$ & $0.66\!\to\!0.54$ \\
\midrule
GSM8K & $\mathrm{conf}_{\text{new}}$ & $0.76\!\to\!0.74$ & $0.65\!\to\!0.61$ & $0.70\!\to\!0.62$ & $0.78\!\to\!0.76$ & $0.62\!\to\!0.57$ \\
 & Likelihood drift (new traj.) & $0.63\!\to\!0.63$ & $0.51\!\to\!0.51$ & $0.52\!\to\!0.51$ & $0.65\!\to\!0.64$ & $0.52\!\to\!0.53$ \\
MATH-full & $\mathrm{conf}_{\text{new}}$ & $0.68\!\to\!0.63$ & $0.62\!\to\!0.58$ & $0.63\!\to\!0.51$ & $0.57\!\to\!0.54$ & $0.67\!\to\!0.55$ \\
 & Likelihood drift (new traj.) & $0.62\!\to\!0.59$ & $0.54\!\to\!0.55$ & $0.60\!\to\!0.59$ & $0.54\!\to\!0.51$ & $0.56\!\to\!0.57$ \\
HumanEval & $\mathrm{conf}_{\text{new}}$ & $0.81\!\to\!0.79$ & $0.75\!\to\!0.68$ & $0.67\!\to\!0.69$ & $0.75\!\to\!0.69$ & $0.61\!\to\!0.55$ \\
 & Likelihood drift (new traj.) & $0.59\!\to\!0.59$ & $0.73\!\to\!0.67$ & $0.54\!\to\!0.57$ & $0.80\!\to\!0.76$ & $0.64\!\to\!0.52$ \\
MBPP & $\mathrm{conf}_{\text{new}}$ & $0.68\!\to\!0.61$ & $0.76\!\to\!0.61$ & $0.81\!\to\!0.60$ & $0.67\!\to\!0.57$ & $0.62\!\to\!0.59$ \\
 & Likelihood drift (new traj.) & $0.75\!\to\!0.66$ & $0.74\!\to\!0.72$ & $0.63\!\to\!0.55$ & $0.75\!\to\!0.63$ & $0.59\!\to\!0.51$ \\
\bottomrule
\end{tabular}
\end{adjustbox}
\caption{\textbf{Signals without label-based filtering.} Each cell reports standalone AUROC on the $M_{\text{old}}$-correct subset $\to$ the full set. New-model confidence provides the common single-model baseline; generation also reports new-trajectory likelihood drift. Appendix Table~\ref{tab:nolabel-full} reports all signals with bootstrap $95\%$ confidence intervals.}
\label{tab:nolabel-summary}
\end{table}

\subsection{Hardness Analysis}
\label{subsec:hardness}
Negative flips are only partly explained by old-model hardness. We operationalize hardness as low old-model confidence on the $M_{\text{old}}$-correct subset, a continuous score rather than a binary label. Flips concentrate among low-confidence examples on MCQ, but this weakens on math reasoning and further on code generation (Figure~\ref{fig:hardness}). On MCQ, old-model margin performs similarly to old-model confidence, while attention entropy and their combinations provide no consistent improvement (Appendix Table~\ref{tab:hardness-mcq-combinations}). The relationship also becomes inconsistent on the Llama-2$\rightarrow$3 boundary case, where the old model is substantially weaker than the new one. Old-model confidence therefore identifies hard examples most consistently within the five in-scope version updates, and adding it as a candidate signal in the added-value test gives only modest gains on some MCQ updates, with no consistent gain across tasks and model updates.

\begin{figure}[!t]
\centering
\includegraphics[width=1\textwidth]{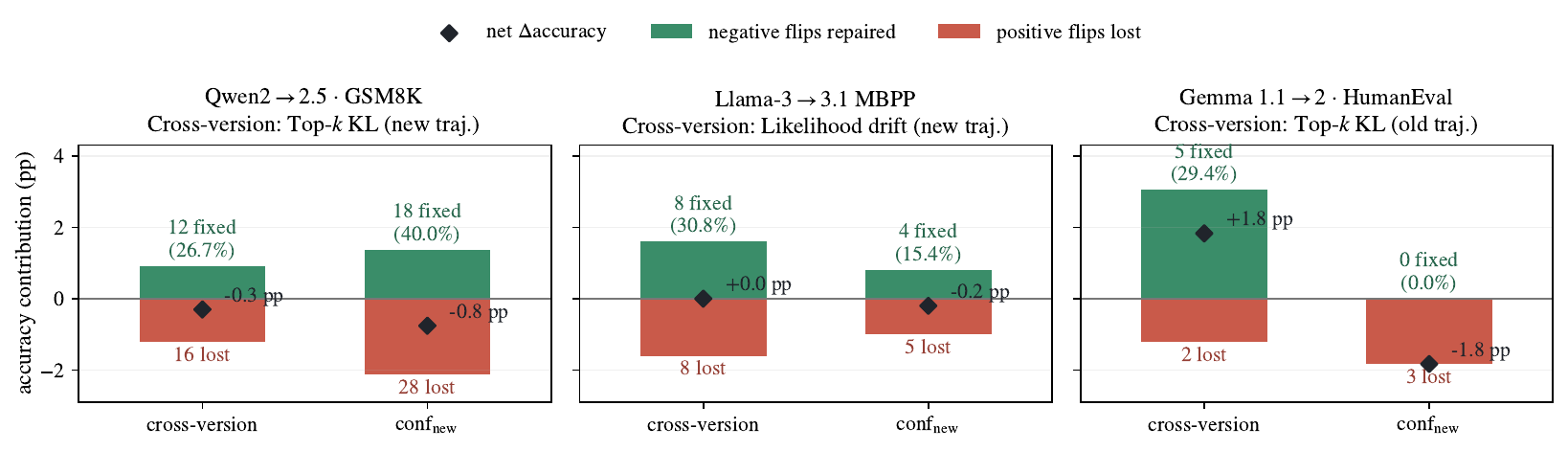}
\caption{\textbf{Selective fallback trade-offs at a fixed 10\% budget.} Green shows negative flips repaired, red shows positive flips lost, and diamonds show the net accuracy change.}
\label{fig:fallback-tradeoff}
\end{figure}

\subsection{Signals Without Label-Based Filtering}
\label{sec:label-free}

The previous analyses use the $M_{\text{old}}$-correct subset, which requires ground-truth labels that are often unavailable in deployment. We therefore repeat the analysis on the full evaluation set without label-based filtering. Table~\ref{tab:nolabel-summary} summarizes the common confidence baseline and new-trajectory likelihood drift, and Appendix Table~\ref{tab:nolabel-full} reports all signals and variants with confidence intervals.

Removing the filter affects both signal types, but the pattern varies by task and update. On MMLU-Pro, new-model confidence drops from $0.82$--$0.85$ on the $M_{\text{old}}$-correct subset to $0.59$--$0.63$ on the full set, with similar drops on GPQA. Some cross-version signals are more stable: new-trajectory likelihood drift decreases only from $0.80$ to $0.76$ on HumanEval for Qwen2$\rightarrow$2.5, from $0.74$ to $0.72$ on MBPP for Llama-3$\rightarrow$3.1, and from $0.62$ to $0.59$ on MATH-full for Qwen2.5$\rightarrow$3. Representation drift remains near random on GSM8K ($\leq0.59$ even on the $M_{\text{old}}$-correct subset). Thus, some cross-version signals remain informative without label-based filtering, although the result is not universal.

\subsection{Selective Fallback}
The full-set analysis suggests ranking requests by regression risk and routing the highest-risk fraction to the old model. We pre-specify a small $10\%$ traffic budget for this proof of concept, rather than selecting the best budget from the test set. Figure~\ref{fig:fallback-tradeoff} shows the resulting trade-off, Appendix Table~\ref{tab:fallback} reports the exact values, and Appendix Figure~\ref{fig:fallback-full-curves} gives the full budget range.

The better trigger depends on the update. On Qwen2$\rightarrow$2.5 GSM8K, new-model confidence catches more flips than the cross-version candidate ($40.0\%$ vs.\ $26.7\%$). The cross-version signal catches more on Llama-3$\rightarrow$3.1 MBPP ($30.8\%$ vs.\ $15.4\%$) and Gemma 1.1$\rightarrow$2 HumanEval ($29.4\%$ vs.\ $0.0\%$), but only the last case gives a clear net gain ($+1.8$ points). Thus, fallback helps only when repaired negative flips outweigh correct new-model answers lost by reverting. Signal selection is retrospective; deployment requires separate calibration or a fixed direction.

%% file: sections/conclusion.tex
\section{Conclusion}

LLM version updates can raise average accuracy while still causing sample-level regression on individual samples. Across five same-family updates, one Llama-2$\rightarrow$3 boundary case, and six benchmarks, we find that no single inference-time signal predicts these regressions: new-model confidence wins on MCQ and simpler math, cross-version likelihood/KL signals win on harder math and code, and old-model confidence alone cannot explain which samples flip. Some cross-version signals remain informative without label-based filtering, supporting a proof-of-concept selective fallback, though combining signals rarely beats the best individual one. Practitioners should therefore match the signal to the task and update rather than defaulting to confidence.

\paragraph{Limitations.} Our task-output-structure account of this pattern is consistent with the results but remains a hypothesis rather than a mechanistic explanation. Future work could extend this framework beyond model version updates to other deployment-time changes, such as agentic scaffolding or system-prompt changes, that can similarly cause sample-level regression.

%% file: sections/appendix.tex
\section{Technical Appendix}
\begin{figure}[ht]
\centering
\includegraphics[width=\textwidth]{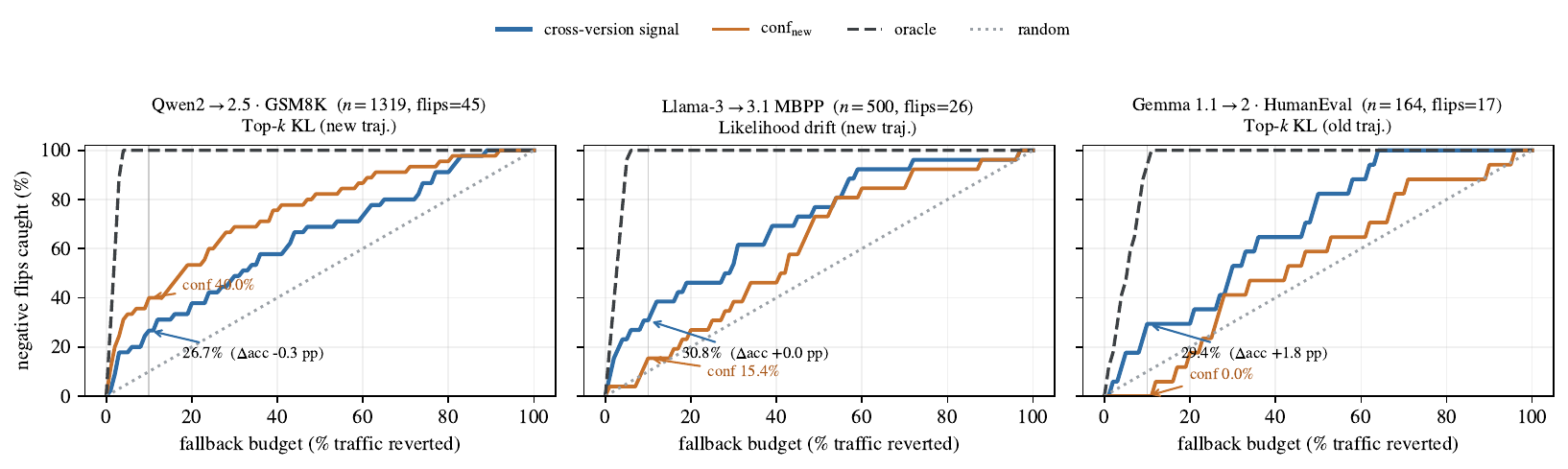}
\includegraphics[width=\textwidth]{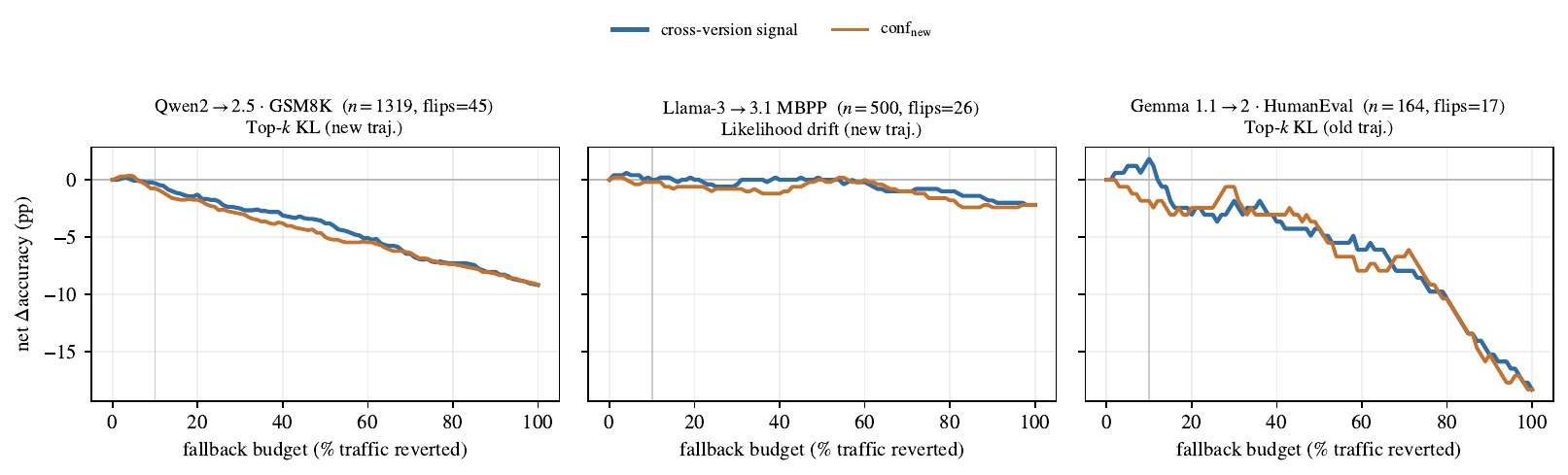}
\caption{\textbf{Selective fallback over the full budget range.} The top row reports negative flips repaired, and the bottom row reports net accuracy change. The vertical line marks the fixed $10\%$ budget used in the main analysis.}
\label{fig:fallback-full-curves}
\end{figure}

This appendix collects the study configuration tables, the expanded per-cell results behind the main-text heatmap, and the additional analyses. Its tables and figures are numbered independently of the main paper. Tables~\ref{tab:pair-metadata}--\ref{tab:benchmarks} give the model-update pairs, the signal definitions, and the benchmark overview. Tables~\ref{tab:mcq-signals}--\ref{tab:gen-code-signals} report baseline values, exact gains, and the old/new model or trajectory variants where applicable; Figure~\ref{fig:signal-heatmaps} shows the variants used in the main comparison. Table~\ref{tab:hardness-mcq-combinations} compares old-model hardness signals on MCQ. Table~\ref{tab:complementarity} gives the complementarity analysis; Table~\ref{tab:fallback} reports the selective-fallback results; Figure~\ref{fig:fallback-full-curves} shows the full budget range and accuracy trade-offs; and Table~\ref{tab:nolabel-full} gives the complete label-free results. A worked negative-flip transcript is given in the case study at the end.

\paragraph{Computing environment.} Experiments used one NVIDIA L40S or H100 GPU per job, 16 CPU cores, and 64\,GB of memory. The software environment used Ubuntu 24.04.4, Python 3.12.4, PyTorch 2.12.0, Transformers 5.8.1, scikit-learn 1.8.0, and SciPy 1.17.0.

\begin{table}[h]
\centering
\footnotesize
\setlength{\tabcolsep}{8pt}
\begin{tabular}{@{}lcccc@{}}
\toprule
\textbf{Hugging Face checkpoint} & \textbf{Params} & \textbf{Release} & \textbf{Hidden} & \textbf{Vocab} \\
\midrule
\multicolumn{5}{@{}l}{\emph{Qwen2.5$\rightarrow$Qwen3 \quad(cross-gen, Acc $+4$~pp)}} \\
Qwen/Qwen2.5-7B-Instruct & 7B & 2024.09 & $3584$ & $152{,}064$ \\
Qwen/Qwen3-8B            & 8B & 2025.04 & $4096$ & $152{,}064$ \\
\addlinespace
\multicolumn{5}{@{}l}{\emph{Llama-3$\rightarrow$Llama-3.1 \quad(minor, Acc $+6$~pp)}} \\
meta-llama/Meta-Llama-3-8B-Instruct & 8B & 2024.04 & $4096$ & $128{,}256$ \\
meta-llama/Llama-3.1-8B-Instruct    & 8B & 2024.07 & $4096$ & $128{,}256$ \\
\addlinespace
\multicolumn{5}{@{}l}{\emph{Mistral v0.2$\rightarrow$Mistral v0.3 \quad(minor, vocab$+$, Acc $+7$~pp)}} \\
mistralai/Mistral-7B-Instruct-v0.2 & 7B & 2023.12 & $4096$ & $32{,}000$ \\
mistralai/Mistral-7B-Instruct-v0.3 & 7B & 2024.05 & $4096$ & $32{,}768$ \\
\addlinespace
\multicolumn{5}{@{}l}{\emph{Qwen2$\rightarrow$Qwen2.5 \quad(minor, Acc $+9$~pp)}} \\
Qwen/Qwen2-7B-Instruct   & 7B & 2024.06 & $3584$ & $152{,}064$ \\
Qwen/Qwen2.5-7B-Instruct & 7B & 2024.09 & $3584$ & $152{,}064$ \\
\addlinespace
\multicolumn{5}{@{}l}{\emph{Gemma 1.1$\rightarrow$Gemma 2 \quad(cross-gen, Acc $+20$~pp)}} \\
google/gemma-1.1-7b-it & 7B & 2024.04 & $3072$ & $256{,}000$ \\
google/gemma-2-9b-it   & 9B & 2024.06 & $3584$ & $256{,}000$ \\
\midrule
\multicolumn{5}{@{}l}{\emph{Llama-2$\rightarrow$Llama-3 \quad(capability rebuild, Acc $+29$~pp)}} \\
meta-llama/Llama-2-7b-chat-hf       & 7B & 2023.07 & $4096$ & $32{,}000$ \\
meta-llama/Meta-Llama-3-8B-Instruct & 8B & 2024.04 & $4096$ & $128{,}256$ \\
\bottomrule
\end{tabular}
\caption{\textbf{Model-update pairs.} Each sub-heading names a pair (old$\rightarrow$new same-family version) with its update type and average accuracy gain $\Delta_{\text{Acc}}=\mathrm{Acc}_{\text{new}}-\mathrm{Acc}_{\text{old}}$ across the three task families; the two rows below give its checkpoints by full Hugging Face repository (\texttt{huggingface.co/<name>}). Hidden size and vocabulary are taken from the model configurations.}
\label{tab:pair-metadata}
\end{table}

\begin{table}[h]
\centering
\footnotesize
\setlength{\tabcolsep}{4pt}
\begin{threeparttable}
\begin{tabular}{@{}lllp{5.5cm}@{}}
\toprule
\textbf{Signal} & \textbf{Category} & \textbf{Mode} & \textbf{Definition} \\
\midrule
Confidence ($\mathrm{conf}_{\text{old}}$, $\mathrm{conf}_{\text{new}}$)
  & single-model  & MCQ + gen & max softmax (MCQ) / mean token log-prob (gen) \\
Logit margin
  & single-model  & MCQ       & top-1 $-$ top-2 logits on option tokens \\
Attention entropy
  & single-model  & MCQ       & mean per-head R\'enyi-2 entropy at the final prompt position \\
\midrule
Output KL/JSD (forward / reverse / JSD)
  & cross-version & MCQ       & $\mathrm{KL}/\mathrm{JSD}$ on option distributions \\
Likelihood drift (old / new traj.)
  & cross-version & gen & per-token NLL difference, computed on the old or new model's generated trajectory \\
Top-$k$ token KL (old / new traj.)
  & cross-version & gen & approximate token-distribution KL on the old or new trajectory \\
Representation drift
  & cross-version & MCQ + gen & $1-\cos$ of last-layer hidden states at the final prompt position (MCQ) or on a shared trajectory (gen) \\
\bottomrule
\end{tabular}

\end{threeparttable}
\caption{\textbf{Signals evaluated.}}
\label{tab:signals}
\end{table}

\begin{table}[h]
\centering
\footnotesize
\setlength{\tabcolsep}{6pt}
\begin{adjustbox}{max width=\textwidth}
\begin{tabular}{@{}llrlll@{}}
\toprule
\textbf{Family} & \textbf{Benchmark} & \textbf{Initial $N$} & \textbf{Output} & \textbf{Grading} & \textbf{Generation cap} \\
\midrule
\multirow{2}{*}{MCQ}
  & MMLU-Pro & $3{,}000$ & one option & option match & --- \\
  & GPQA Main & $448$ & one option & option match & --- \\
\midrule
\multirow{2}{*}{Math reasoning}
  & GSM8K & $1{,}319$ & reasoning + final answer & normalized answer match & $1{,}024$ tokens \\
  & MATH-full & $5{,}000$ & reasoning + final answer & normalized answer match & $2{,}048$ tokens \\
\midrule
\multirow{2}{*}{Code generation}
  & HumanEval & $164$ & program & unit tests & $512$ tokens \\
  & MBPP & $500$ & program & unit tests & $256$--$320$ tokens \\
\bottomrule
\end{tabular}
\end{adjustbox}
\caption{\textbf{Benchmark overview.} Initial dataset sizes, output formats, grading rules, and decoding limits. Generations that reach the token cap are excluded, so the final analysis size can be smaller.}
\label{tab:benchmarks}
\end{table}

\begin{table}[h]
\centering
\footnotesize
\setlength{\tabcolsep}{4pt}
\begin{tabular}{@{}lccccc@{}}
\toprule
\textbf{Signal}
  & \textbf{Qwen2.5$\rightarrow$3}
  & \textbf{Llama-3$\rightarrow$3.1}
  & \textbf{Mistral v0.2$\rightarrow$v0.3}
  & \textbf{Qwen2$\rightarrow$2.5}
  & \textbf{Gemma 1.1$\rightarrow$2} \\
\midrule
\multicolumn{6}{@{}l}{\emph{MMLU-Pro ($N{=}3{,}000$ per pair)}} \\
Accuracy gain (pp)          & $+2.1$ & $-0.6$ & $+1.8$ & $+8.2$ & $+14.2$ \\
$\mathrm{conf}_{\text{new}}$ (baseline CV-AUROC)
  & $0.85$  & $0.83$  & $0.82$  & $0.83$ & $0.83$ \\
$\mathrm{conf}_{\text{old}}$     & $+0.013$ & $+0.036$ & $+0.016$ & $-0.014$ & $+0.001$ \\
Logit margin (old)          & $-0.007$ & $+0.023$ & $-0.002$ & $-0.025$ & $-0.048$ \\
Logit margin (new)          & $+0.001$ & $-0.010$ & $-0.003$ & $-0.003$ & $-0.002$ \\
Attention entropy (old)     & $-0.051$ & $+0.000$ & $-0.022$ & $-0.067$ & $-0.064$ \\
Attention entropy (new)     & $-0.065$ & $+0.001$ & $-0.034$ & $-0.061$ & $-0.070$ \\
Output KL (JSD)             & $+0.146$ & $+0.126$ & $+0.177$ & $+0.161$ & $+0.163$ \\
Representation drift        & ---     & $+0.009$ & $+0.041$ & $-0.046$ & --- \\
$n_{\text{old-correct}}$ & $1{,}245$ & $951$ & $826$ & $999$ & $880$ \\
Negative flips           & $365$     & $277$     & $234$ & $274$     & $217$ \\
Flip rate (NFR, \%)      & $29.3$ & $29.1$ & $28.3$ & $27.4$ & $24.7$ \\
\midrule
\multicolumn{6}{@{}l}{\emph{GPQA \texttt{gpqa\_main} ($N{=}448$ per pair)}} \\
Accuracy gain (pp)          & $+4.5$ & $+1.1$ & $+1.3$ & $+2.5$ & $+7.6$ \\
$\mathrm{conf}_{\text{new}}$ (baseline CV-AUROC)
  & $0.74$  & $0.66$  & $0.67$  & $0.70$ & $0.65$ \\
$\mathrm{conf}_{\text{old}}$     & $+0.029$ & $+0.056$ & $-0.012$ & $+0.051$ & $-0.008$ \\
Logit margin (old)          & $+0.007$ & $+0.024$ & $+0.040$ & $+0.050$ & $+0.008$ \\
Logit margin (new)          & $-0.003$ & $-0.002$ & $-0.005$ & $+0.004$ & $-0.006$ \\
Attention entropy (old)     & $-0.028$ & $-0.023$ & $-0.013$ & $-0.048$ & $-0.053$ \\
Attention entropy (new)     & $+0.027$ & $+0.006$ & $-0.049$ & $-0.039$ & $-0.039$ \\
Output KL (JSD)             & $+0.243$ & $+0.187$ & $+0.313$ & $+0.302$ & $+0.334$ \\
Representation drift        & ---     & $-0.008$ & $+0.004$ & $-0.020$ & --- \\
$n_{\text{old-correct}}$ & $147$ & $136$ & $126$ & $136$ & $127$ \\
Negative flips           & $61$  & $40$  & $38$ & $61$ & $60$ \\
Flip rate (NFR, \%)      & $41.5$ & $29.4$ & $30.2$ & $44.9$ & $47.2$ \\
\bottomrule
\end{tabular}
\caption{\textbf{MCQ-phase results.} The baseline row reports CV-AUROC using $\mathrm{conf}_{\text{new}}$ alone; all signal rows report the change in CV-AUROC ($\Delta$) after adding that signal to the baseline. The accuracy gain row is in percentage points (pp), $\Delta_{\text{Acc}}=\mathrm{Acc}_{\text{new}}-\mathrm{Acc}_{\text{old}}$, and this convention is used in all signal tables. Output KL and JSD are reported separately as direct measures of output-distribution change. Logit margin and attention entropy are reported for both model versions; the main comparison uses the old-model variants, while the new-model variants largely overlap with the $\mathrm{conf}_{\text{new}}$ baseline. NFR is the fraction of the $M_{\text{old}}$-correct subset that becomes incorrect under $M_{\text{new}}$.}
\label{tab:mcq-signals}
\end{table}

\begin{table}[h]
\centering
\footnotesize
\setlength{\tabcolsep}{4pt}
\begin{tabular}{@{}lccccc@{}}
\toprule
\textbf{Signal}
  & \textbf{Qwen2.5$\rightarrow$3}
  & \textbf{Llama-3$\rightarrow$3.1}
  & \textbf{Mistral v0.2$\rightarrow$v0.3}
  & \textbf{Qwen2$\rightarrow$2.5}
  & \textbf{Gemma 1.1$\rightarrow$2} \\
\midrule
\multicolumn{6}{@{}l}{\emph{GSM8K ($N{=}1{,}319$ per pair)}} \\
Accuracy gain (pp)                & $+3.1$ & $+4.8$ & $+12.7$ & $+9.2$ & $+35.6$ \\
$\mathrm{conf}_{\text{new}}$ (baseline CV-AUROC)
  & $0.76$ & $0.65$ & $0.69$ & $0.78$ & $0.62$ \\
$\mathrm{conf}_{\text{old}}$     & $-0.000$ & $+0.008$ & $-0.004$ & $-0.009$ & $-0.024$ \\
Likelihood drift (new traj.)      & $-.005$ & $-.007$ & $+.037$ & $+.003$ & $-.065$ \\
Top-$k$ token KL (new traj.)      & $-.001$ & $-.013$ & $+.038$ & $-.002$ & $-.071$ \\
Likelihood drift (old traj.)      & $+.019$ & $-.014$ & $-.002$ & $+.004$ & $-.046$ \\
Top-$k$ token KL (old traj.)      & $+.016$ & $-.009$ & $-.002$ & $+.013$ & $-.032$ \\
Rep.\ drift, last tok (new traj.) & ---     & $-.009$ & $-.002$ & $+.009$ & --- \\
$n_{\text{old-correct}}$ & $1{,}196$ & $1{,}047$ & $559$ & $1{,}075$ & $657$ \\
Negative flips           & $31$ & $85$ & $132$ & $45$ & $32$ \\
Flip rate (NFR, \%)      & $2.6$ & $8.1$ & $23.6$ & $4.2$ & $4.9$ \\
\midrule
\multicolumn{6}{@{}l}{\emph{MATH-full ($N{=}5{,}000$ per pair)}} \\
Accuracy gain (pp)                & $+7.5$ & $+23.6$ & $+1.8$ & $+19.6$ & $+32.0$ \\
$\mathrm{conf}_{\text{new}}$ (baseline CV-AUROC)
  & $0.68$ & $0.61$ & $0.63$ & $0.57$ & $0.67$ \\
$\mathrm{conf}_{\text{old}}$     & $+0.008$ & $+0.036$ & $+0.001$ & $-0.012$ & $+0.039$ \\
Likelihood drift (new traj.)      & $+.079$ & $+.007$ & $+.059$ & $-.007$ & $+.055$ \\
Top-$k$ token KL (new traj.)      & $+.086$ & $+.001$ & $+.058$ & $-.007$ & $+.063$ \\
Likelihood drift (old traj.)      & $+.026$ & $+.003$ & $+.035$ & $-.024$ & $-.001$ \\
Top-$k$ token KL (old traj.)      & $+.029$ & $+.000$ & $+.036$ & $-.027$ & $+.000$ \\
Rep.\ drift, last tok (new traj.) & ---     & $+.025$ & $-.006$ & $+.013$ & --- \\
$n_{\text{old-correct}}$ & $3{,}437$ & $1{,}139$ & $449$ & $2{,}510$ & $650$ \\
Negative flips           & $180$ & $228$ & $248$ & $152$ & $115$ \\
Flip rate (NFR, \%)      & $5.2$ & $20.0$ & $55.2$ & $6.1$ & $17.7$ \\
\bottomrule
\end{tabular}
\caption{\textbf{Math-phase results.} The baseline row reports CV-AUROC using $\mathrm{conf}_{\text{new}}$ alone; all signal rows report the change in CV-AUROC ($\Delta$) after adding that signal to the baseline. Each cross-version signal comes in an old-trajectory and a new-trajectory variant; Figure~\ref{fig:signal-heatmaps} shows the new-trajectory variants used in the main comparison. NFR is the fraction of the $M_{\text{old}}$-correct subset that becomes incorrect under $M_{\text{new}}$.}
\label{tab:gen-math-signals}
\end{table}

\begin{table}[h]
\centering
\footnotesize
\setlength{\tabcolsep}{4pt}
\begin{tabular}{@{}lccccc@{}}
\toprule
\textbf{Signal}
  & \textbf{Qwen2.5$\rightarrow$3}
  & \textbf{Llama-3$\rightarrow$3.1}
  & \textbf{Mistral v0.2$\rightarrow$v0.3}
  & \textbf{Qwen2$\rightarrow$2.5}
  & \textbf{Gemma 1.1$\rightarrow$2} \\
\midrule
\multicolumn{6}{@{}l}{\emph{HumanEval ($N{=}164$ per pair)}} \\
Accuracy gain (pp)                    & $+1.2$ & $+6.1$ & $+13.0$ & $+6.1$ & $+18.3$ \\
$\mathrm{conf}_{\text{new}}$ (baseline CV-AUROC) & $0.80$ & $0.76$ & $0.70$ & $0.75$ & $0.42$ \\
$\mathrm{conf}_{\text{old}}$          & $-.003$ & $-.009$ & $-.160$ & $-.009$ & $-.060$ \\
Likelihood drift (new traj.)          & $-.012$ & $+.035$ & $-.112$ & $+.042$ & $+.215$ \\
Top-$k$ token KL (new traj.)          & $-.018$ & $+.023$ & $-.079$ & $+.045$ & $+.232$ \\
Likelihood drift (old traj.)          & $-.039$ & $+.012$ & $-.157$ & $-.015$ & $+.303$ \\
Top-$k$ token KL (old traj.)          & $-.048$ & $+.021$ & $-.143$ & $-.031$ & $+.337$ \\
Rep.\ drift, last tok (new traj.)     & --- & $-.063$ & $-.138$ & $-.010$ & --- \\
$n_{\text{old-correct}}$              & $125$ & $86$ & $43$ & $115$ & $67$ \\
Negative flips                        & $15$ & $17$ & $10$ & $12$ & $17$ \\
Flip rate (NFR, \%)                   & $12.0$ & $19.8$ & $23.3$ & $10.4$ & $25.4$ \\
\midrule
\multicolumn{6}{@{}l}{\emph{MBPP ($N{=}500$ per pair)}} \\
Accuracy gain (pp)                    & $+2.8$ & $+2.2$ & $+9.4$ & $+8.5$ & $+15.0$ \\
$\mathrm{conf}_{\text{new}}$ (baseline CV-AUROC) & $0.69$ & $0.76$ & $0.79$ & $0.66$ & $0.62$ \\
$\mathrm{conf}_{\text{old}}$          & $+.014$ & $-.012$ & $-.035$ & $+.042$ & $+.003$ \\
Likelihood drift (new traj.)          & $+.069$ & $+.020$ & $-.023$ & $+.103$ & $+.075$ \\
Top-$k$ token KL (new traj.)          & $+.067$ & $-.009$ & $-.020$ & $+.112$ & $+.091$ \\
Likelihood drift (old traj.)          & $+.037$ & $-.007$ & $-.023$ & $+.124$ & $-.016$ \\
Top-$k$ token KL (old traj.)          & $+.042$ & $-.004$ & $-.026$ & $+.025$ & $-.016$ \\
Rep.\ drift, last tok (new traj.)     & --- & $-.020$ & $-.023$ & $-.041$ & --- \\
$n_{\text{old-correct}}$              & $296$ & $257$ & $121$ & $248$ & $186$ \\
Negative flips                        & $40$ & $26$ & $17$ & $29$ & $24$ \\
Flip rate (NFR, \%)                   & $13.5$ & $10.1$ & $14.0$ & $11.7$ & $12.9$ \\
\bottomrule
\end{tabular}
\caption{\textbf{Code-phase results.} The baseline row reports CV-AUROC using $\mathrm{conf}_{\text{new}}$ alone; all signal rows report the change in CV-AUROC ($\Delta$) after adding that signal to the baseline. Each cross-version signal comes in an old-trajectory and a new-trajectory variant; Figure~\ref{fig:signal-heatmaps} shows the new-trajectory variants used in the main comparison. NFR is the fraction of the $M_{\text{old}}$-correct subset that becomes incorrect under $M_{\text{new}}$.}
\label{tab:gen-code-signals}
\end{table}

\begin{table}[t]
\centering
\footnotesize
\setlength{\tabcolsep}{4pt}
\begin{tabular}{@{}lccccc@{}}
\toprule
\textbf{Old-model signal(s)}
  & \textbf{Qwen2.5$\rightarrow$3}
  & \textbf{Llama-3$\rightarrow$3.1}
  & \textbf{Mistral v0.2$\rightarrow$v0.3}
  & \textbf{Qwen2$\rightarrow$2.5}
  & \textbf{Gemma 1.1$\rightarrow$2} \\
\midrule
\multicolumn{6}{@{}l}{\emph{MMLU-Pro}} \\
$\mathrm{conf}_{\text{old}}$                    & $0.810$ & $0.840$ & $0.755$ & $0.765$ & $0.737$ \\
Logit margin                                    & $0.809$ & $0.828$ & $0.765$ & $0.760$ & $0.735$ \\
Attention entropy                               & $0.650$ & $0.581$ & $0.626$ & $0.589$ & $0.636$ \\
$\mathrm{conf}_{\text{old}}$ + margin           & $0.810$ & $0.837$ & $0.766$ & $0.760$ & $0.739$ \\
$\mathrm{conf}_{\text{old}}$ + entropy          & $0.756$ & $0.837$ & $0.683$ & $0.752$ & $0.702$ \\
Margin + entropy                                & $0.815$ & $0.834$ & $0.765$ & $0.758$ & $0.744$ \\
$\mathrm{conf}_{\text{old}}$ + margin + entropy & $0.816$ & $0.843$ & $0.765$ & $0.758$ & $0.743$ \\
\midrule
\multicolumn{6}{@{}l}{\emph{GPQA}} \\
$\mathrm{conf}_{\text{old}}$                    & $0.720$ & $0.727$ & $0.575$ & $0.745$ & $0.677$ \\
Logit margin                                    & $0.707$ & $0.703$ & $0.697$ & $0.735$ & $0.667$ \\
Attention entropy                               & $0.550$ & $0.563$ & $0.591$ & $0.529$ & $0.506$ \\
$\mathrm{conf}_{\text{old}}$ + margin           & $0.718$ & $0.725$ & $0.725$ & $0.741$ & $0.602$ \\
$\mathrm{conf}_{\text{old}}$ + entropy          & $0.681$ & $0.720$ & $0.575$ & $0.744$ & $0.581$ \\
Margin + entropy                                & $0.708$ & $0.682$ & $0.700$ & $0.733$ & $0.672$ \\
$\mathrm{conf}_{\text{old}}$ + margin + entropy & $0.716$ & $0.718$ & $0.728$ & $0.743$ & $0.622$ \\
\bottomrule
\end{tabular}
\caption{\textbf{Old-model signals for identifying hard MCQ examples.} Values are absolute 5-fold CV-AUROC on the $M_{\text{old}}$-correct subset. All rows use logistic regression fitted within each training fold; they therefore differ from the unfitted standalone AUROC in Figure~\ref{fig:hardness}.}
\label{tab:hardness-mcq-combinations}
\end{table}

\begin{table}[h]
\centering
\footnotesize
\setlength{\tabcolsep}{4pt}
\begin{tabular}{@{}lllccc@{}}
\toprule
\textbf{Bench} & \textbf{Pair} & \textbf{Best single signal}
  & \textbf{Best $\Delta$} & \textbf{All-cross $\Delta$} & \textbf{Gain} \\
\midrule
\multicolumn{6}{@{}l}{\emph{MCQ: MMLU-Pro}} \\
MMLU-Pro & Qwen2.5$\rightarrow$3 & Output KL (JSD) & $+0.146$ & $+0.146$ & $+0.000$ \\
MMLU-Pro & Llama-3$\rightarrow$3.1 & Output KL (JSD) & $+0.126$ & $+0.131$ & $+0.005$ \\
MMLU-Pro & Mistral v0.2$\rightarrow$v0.3 & Output KL (JSD) & $+0.177$ & $+0.177$ & $-0.001$ \\
MMLU-Pro & Qwen2$\rightarrow$2.5 & Output KL (JSD) & $+0.161$ & $+0.162$ & $+0.001$ \\
MMLU-Pro & Gemma 1.1$\rightarrow$2 & Output KL (JSD) & $+0.163$ & $+0.163$ & $-0.000$ \\
\midrule
\multicolumn{6}{@{}l}{\emph{MCQ: GPQA}} \\
GPQA & Qwen2.5$\rightarrow$3 & Output KL (JSD) & $+0.243$ & $+0.243$ & $+0.001$ \\
GPQA & Llama-3$\rightarrow$3.1 & Output KL (fwd) & $+0.212$ & $+0.215$ & $+0.003$ \\
GPQA & Mistral v0.2$\rightarrow$v0.3 & Output KL (JSD) & $+0.313$ & $+0.312$ & $-0.002$ \\
GPQA & Qwen2$\rightarrow$2.5 & Output KL (JSD) & $+0.302$ & $+0.298$ & $-0.004$ \\
GPQA & Gemma 1.1$\rightarrow$2 & Output KL (JSD) & $+0.334$ & $+0.334$ & $+0.000$ \\
\midrule
\multicolumn{6}{@{}l}{\emph{Math: GSM8K}} \\
GSM8K & Qwen2.5$\rightarrow$3 & Likelihood drift (old traj.) & $+0.019$ & $+0.010$ & $-0.009$ \\
GSM8K & Llama-3$\rightarrow$3.1 & Rep.\ drift, mean (new traj.) & $-0.004$ & $-0.007$ & $-0.003$ \\
GSM8K & Mistral v0.2$\rightarrow$v0.3 & Top-$k$ token KL (new traj.) & $+0.038$ & $+0.038$ & $+0.000$ \\
GSM8K & Qwen2$\rightarrow$2.5 & Rep.\ drift, mean (new traj.) & $+0.029$ & $+0.034$ & $+0.005$ \\
GSM8K & Gemma 1.1$\rightarrow$2 & Top-$k$ token KL (old traj.) & $-0.032$ & $-0.102$ & $-0.069$ \\
\midrule
\multicolumn{6}{@{}l}{\emph{Math: full MATH set}} \\
MATH-full & Qwen2.5$\rightarrow$3 & Top-$k$ token KL (new traj.) & $+0.086$ & $+0.083$ & $-0.003$ \\
MATH-full & Llama-3$\rightarrow$3.1 & Rep.\ drift, last tok (new traj.) & $+0.025$ & $+0.033$ & $+0.008$ \\
MATH-full & Mistral v0.2$\rightarrow$v0.3 & Likelihood drift (new traj.) & $+0.059$ & $+0.073$ & $+0.013$ \\
MATH-full & Qwen2$\rightarrow$2.5 & Rep.\ drift, mean (new traj.) & $+0.135$ & $+0.130$ & $-0.005$ \\
MATH-full & Gemma 1.1$\rightarrow$2 & Top-$k$ token KL (new traj.) & $+0.063$ & $+0.051$ & $-0.012$ \\
\midrule
\multicolumn{6}{@{}l}{\emph{Code: HumanEval}} \\
HumanEval & Qwen2.5$\rightarrow$3 & Likelihood drift (new traj.) & $-0.012$ & $-0.009$ & $+0.003$ \\
HumanEval & Llama-3$\rightarrow$3.1 & Likelihood drift (new traj.) & $+0.035$ & $+0.101$ & $+0.066$ \\
HumanEval & Mistral v0.2$\rightarrow$v0.3 & Rep.\ drift, mean (new traj.) & $-0.017$ & $-0.214$ & $-0.198$ \\
HumanEval & Qwen2$\rightarrow$2.5 & Top-$k$ token KL (new traj.) & $+0.045$ & $-0.017$ & $-0.062$ \\
HumanEval & Gemma 1.1$\rightarrow$2 & Top-$k$ token KL (old traj.) & $+0.337$ & $+0.295$ & $-0.042$ \\
\midrule
\multicolumn{6}{@{}l}{\emph{Code: MBPP}} \\
MBPP & Qwen2.5$\rightarrow$3 & Likelihood drift (new traj.) & $+0.069$ & $+0.124$ & $+0.054$ \\
MBPP & Llama-3$\rightarrow$3.1 & Likelihood drift (new traj.) & $+0.020$ & $+0.043$ & $+0.023$ \\
MBPP & Mistral v0.2$\rightarrow$v0.3 & Top-$k$ token KL (new traj.) & $-0.020$ & $-0.023$ & $-0.002$ \\
MBPP & Qwen2$\rightarrow$2.5 & Likelihood drift (old traj.) & $+0.124$ & $+0.124$ & $+0.001$ \\
MBPP & Gemma 1.1$\rightarrow$2 & Top-$k$ token KL (new traj.) & $+0.091$ & $+0.074$ & $-0.017$ \\
\midrule
\multicolumn{6}{@{}l}{\emph{Out of scope: capability rebuild}} \\
MMLU-Pro & Llama-2$\rightarrow$3 & Output KL (rev) & $+0.110$ & $+0.112$ & $+0.002$ \\
GPQA & Llama-2$\rightarrow$3 & Output KL (JSD) & $+0.369$ & $+0.359$ & $-0.010$ \\
GSM8K & Llama-2$\rightarrow$3 & Likelihood drift (new traj.) & $+0.039$ & $-0.016$ & $-0.055$ \\
MATH-full & Llama-2$\rightarrow$3 & Top-$k$ token KL (old traj.) & $+0.111$ & $+0.085$ & $-0.026$ \\
HumanEval & Llama-2$\rightarrow$3 & --- (no combo) & --- & --- & --- \\
MBPP & Llama-2$\rightarrow$3 & Likelihood drift (new traj.) & $+0.061$ & $+0.037$ & $-0.025$ \\
\bottomrule
\end{tabular}
\caption{\textbf{Signal complementarity.} Added-value CV-AUROC ($\Delta$) of the best single cross-version signal and the combined cross-version model over $\mathrm{conf}_{\text{new}}$. The out-of-scope Llama-2$\rightarrow$3 rows use unrelated tokenizers, so their token-level cross-version values (likelihood drift, top-$k$ token KL) are not directly comparable and are shown only for completeness.}
\label{tab:complementarity}
\end{table}

\begin{table}[h]
\centering
\footnotesize
\setlength{\tabcolsep}{4pt}
\begin{tabular}{@{}lllcccc@{}}
\toprule
\textbf{Pair} & \textbf{Task} & \textbf{Cross-version signal}
  & \textbf{Flips caught} & \textbf{$\Delta_{\text{Acc}}$ (pp)}
  & \textbf{$\mathrm{conf}_{\text{new}}$} & \textbf{random} \\
\midrule
Qwen2$\rightarrow$2.5     & GSM8K     & Top-$k$ token KL (new traj.) & $26.7\%$ & $-0.3$ & $\mathbf{40.0\%}$ & $10.0\%$ \\
Llama-3$\rightarrow$3.1       & MBPP      & Likelihood drift (new traj.) & $30.8\%$ & $+0.0$ & $15.4\%$ & $10.0\%$ \\
Gemma 1.1$\rightarrow$2 & HumanEval & Top-$k$ token KL (old traj.) & $\mathbf{29.4\%}$ & $+1.8$ & $0.0\%$ & $9.8\%$ \\
\bottomrule
\end{tabular}
\caption{\textbf{Selective fallback with a $10\%$ budget.} Flips caught is the percentage of repaired negative flips, and $\Delta_{\text{Acc}}$ is the accuracy change in percentage points relative to always serving $M_{\text{new}}$. The $\mathrm{conf}_{\text{new}}$ column uses the same budget, and the random column reports the expected capture from uniformly routing the same rounded number of inputs. Confidence is the better trigger on Qwen2$\rightarrow$2.5 GSM8K, while the cross-version signal only clearly outperforms confidence on Gemma 1.1$\rightarrow$2 HumanEval, where confidence is unreliable (AUROC $0.42$).}
\label{tab:fallback}
\end{table}

\clearpage

{\small
\setlength{\tabcolsep}{5pt}
\setlength{\LTcapwidth}{\textwidth}
\begin{longtable}{@{}llccc@{}}
\caption{\textbf{Label-free per-cell AUROC.} Standalone AUROC of all formally evaluated signals and their main old/new model or trajectory variants on the $M_{\text{old}}$-correct subset and the full evaluation set, with bootstrap $95\%$ confidence intervals. Entries marked --- are unavailable because representation drift requires equal hidden size, which holds only for three update pairs: Llama-3$\rightarrow$3.1, Mistral v0.2$\rightarrow$v0.3, and Qwen2$\rightarrow$2.5.}
\label{tab:nolabel-full} \\
\toprule
\textbf{Pair} & \textbf{Task} & \textbf{Signal} & \textbf{$M_{\text{old}}$-correct} & \textbf{Full} \\
\midrule
\endfirsthead
\multicolumn{5}{@{}l}{\emph{Table~\ref{tab:nolabel-full}, continued from previous page}} \\
\toprule
\textbf{Pair} & \textbf{Task} & \textbf{Signal} & \textbf{$M_{\text{old}}$-correct} & \textbf{Full} \\
\midrule
\endhead
\midrule
\multicolumn{5}{r@{}}{\emph{continued on next page}} \\
\endfoot
\bottomrule
\endlastfoot
Qwen2.5$\rightarrow$3 & MMLU-Pro & $\mathrm{conf}_{\text{old}}$ & $0.81\,[0.78,0.83]$ & $0.59\,[0.56,0.62]$ \\
Llama-3$\rightarrow$3.1 & MMLU-Pro & $\mathrm{conf}_{\text{old}}$ & $0.84\,[0.81,0.86]$ & $0.57\,[0.54,0.60]$ \\
Mistral v0.2$\rightarrow$v0.3 & MMLU-Pro & $\mathrm{conf}_{\text{old}}$ & $0.75\,[0.72,0.79]$ & $0.55\,[0.52,0.59]$ \\
Qwen2$\rightarrow$2.5 & MMLU-Pro & $\mathrm{conf}_{\text{old}}$ & $0.76\,[0.73,0.80]$ & $0.53\,[0.50,0.56]$ \\
Gemma 1.1$\rightarrow$2 & MMLU-Pro & $\mathrm{conf}_{\text{old}}$ & $0.74\,[0.70,0.77]$ & $0.54\,[0.50,0.58]$ \\
Qwen2.5$\rightarrow$3 & MMLU-Pro & $\mathrm{conf}_{\text{new}}$ & $0.85\,[0.83,0.87]$ & $0.63\,[0.61,0.66]$ \\
Llama-3$\rightarrow$3.1 & MMLU-Pro & $\mathrm{conf}_{\text{new}}$ & $0.83\,[0.80,0.85]$ & $0.60\,[0.58,0.63]$ \\
Mistral v0.2$\rightarrow$v0.3 & MMLU-Pro & $\mathrm{conf}_{\text{new}}$ & $0.82\,[0.79,0.85]$ & $0.59\,[0.56,0.63]$ \\
Qwen2$\rightarrow$2.5 & MMLU-Pro & $\mathrm{conf}_{\text{new}}$ & $0.83\,[0.80,0.85]$ & $0.61\,[0.59,0.64]$ \\
Gemma 1.1$\rightarrow$2 & MMLU-Pro & $\mathrm{conf}_{\text{new}}$ & $0.83\,[0.81,0.86]$ & $0.62\,[0.59,0.65]$ \\
Qwen2.5$\rightarrow$3 & MMLU-Pro & Logit margin (old) & $0.81\,[0.78,0.83]$ & $0.59\,[0.56,0.62]$ \\
Llama-3$\rightarrow$3.1 & MMLU-Pro & Logit margin (old) & $0.83\,[0.80,0.85]$ & $0.58\,[0.55,0.61]$ \\
Mistral v0.2$\rightarrow$v0.3 & MMLU-Pro & Logit margin (old) & $0.76\,[0.73,0.80]$ & $0.56\,[0.52,0.59]$ \\
Qwen2$\rightarrow$2.5 & MMLU-Pro & Logit margin (old) & $0.76\,[0.73,0.79]$ & $0.54\,[0.51,0.58]$ \\
Gemma 1.1$\rightarrow$2 & MMLU-Pro & Logit margin (old) & $0.73\,[0.70,0.77]$ & $0.55\,[0.51,0.59]$ \\
Qwen2.5$\rightarrow$3 & MMLU-Pro & Logit margin (new) & $0.85\,[0.83,0.87]$ & $0.64\,[0.61,0.66]$ \\
Llama-3$\rightarrow$3.1 & MMLU-Pro & Logit margin (new) & $0.80\,[0.78,0.83]$ & $0.61\,[0.58,0.64]$ \\
Mistral v0.2$\rightarrow$v0.3 & MMLU-Pro & Logit margin (new) & $0.81\,[0.78,0.84]$ & $0.59\,[0.56,0.63]$ \\
Qwen2$\rightarrow$2.5 & MMLU-Pro & Logit margin (new) & $0.82\,[0.79,0.85]$ & $0.62\,[0.59,0.65]$ \\
Gemma 1.1$\rightarrow$2 & MMLU-Pro & Logit margin (new) & $0.84\,[0.81,0.86]$ & $0.63\,[0.59,0.66]$ \\
Qwen2.5$\rightarrow$3 & MMLU-Pro & Attention entropy (old) & $0.65\,[0.62,0.68]$ & $0.53\,[0.50,0.56]$ \\
Llama-3$\rightarrow$3.1 & MMLU-Pro & Attention entropy (old) & $0.58\,[0.54,0.62]$ & $0.51\,[0.47,0.54]$ \\
Mistral v0.2$\rightarrow$v0.3 & MMLU-Pro & Attention entropy (old) & $0.63\,[0.58,0.67]$ & $0.50\,[0.47,0.54]$ \\
Qwen2$\rightarrow$2.5 & MMLU-Pro & Attention entropy (old) & $0.59\,[0.55,0.63]$ & $0.51\,[0.47,0.54]$ \\
Gemma 1.1$\rightarrow$2 & MMLU-Pro & Attention entropy (old) & $0.64\,[0.60,0.68]$ & $0.55\,[0.51,0.58]$ \\
Qwen2.5$\rightarrow$3 & MMLU-Pro & Attention entropy (new) & $0.64\,[0.61,0.68]$ & $0.52\,[0.49,0.55]$ \\
Llama-3$\rightarrow$3.1 & MMLU-Pro & Attention entropy (new) & $0.57\,[0.53,0.61]$ & $0.51\,[0.48,0.55]$ \\
Mistral v0.2$\rightarrow$v0.3 & MMLU-Pro & Attention entropy (new) & $0.56\,[0.52,0.61]$ & $0.51\,[0.47,0.55]$ \\
Qwen2$\rightarrow$2.5 & MMLU-Pro & Attention entropy (new) & $0.63\,[0.59,0.67]$ & $0.54\,[0.51,0.58]$ \\
Gemma 1.1$\rightarrow$2 & MMLU-Pro & Attention entropy (new) & $0.53\,[0.48,0.58]$ & $0.50\,[0.46,0.55]$ \\
Qwen2.5$\rightarrow$3 & MMLU-Pro & Output KL (old$\Vert$new) & $0.97\,[0.96,0.98]$ & $0.71\,[0.69,0.73]$ \\
Llama-3$\rightarrow$3.1 & MMLU-Pro & Output KL (old$\Vert$new) & $0.93\,[0.91,0.94]$ & $0.74\,[0.72,0.77]$ \\
Mistral v0.2$\rightarrow$v0.3 & MMLU-Pro & Output KL (old$\Vert$new) & $0.99\,[0.98,0.99]$ & $0.75\,[0.73,0.78]$ \\
Qwen2$\rightarrow$2.5 & MMLU-Pro & Output KL (old$\Vert$new) & $0.90\,[0.88,0.92]$ & $0.64\,[0.61,0.66]$ \\
Gemma 1.1$\rightarrow$2 & MMLU-Pro & Output KL (old$\Vert$new) & $0.99\,[0.98,0.99]$ & $0.65\,[0.62,0.67]$ \\
Qwen2.5$\rightarrow$3 & MMLU-Pro & Output KL (new$\Vert$old) & $0.98\,[0.97,0.99]$ & $0.76\,[0.74,0.78]$ \\
Llama-3$\rightarrow$3.1 & MMLU-Pro & Output KL (new$\Vert$old) & $0.91\,[0.89,0.92]$ & $0.74\,[0.71,0.77]$ \\
Mistral v0.2$\rightarrow$v0.3 & MMLU-Pro & Output KL (new$\Vert$old) & $0.98\,[0.97,0.98]$ & $0.76\,[0.73,0.78]$ \\
Qwen2$\rightarrow$2.5 & MMLU-Pro & Output KL (new$\Vert$old) & $0.98\,[0.97,0.98]$ & $0.74\,[0.71,0.76]$ \\
Gemma 1.1$\rightarrow$2 & MMLU-Pro & Output KL (new$\Vert$old) & $0.99\,[0.98,0.99]$ & $0.71\,[0.69,0.74]$ \\
Qwen2.5$\rightarrow$3 & MMLU-Pro & Output JSD & $1.00\,[1.00,1.00]$ & $0.75\,[0.74,0.77]$ \\
Llama-3$\rightarrow$3.1 & MMLU-Pro & Output JSD & $0.94\,[0.92,0.95]$ & $0.75\,[0.73,0.78]$ \\
Mistral v0.2$\rightarrow$v0.3 & MMLU-Pro & Output JSD & $1.00\,[0.99,1.00]$ & $0.77\,[0.75,0.79]$ \\
Qwen2$\rightarrow$2.5 & MMLU-Pro & Output JSD & $0.98\,[0.97,0.99]$ & $0.72\,[0.70,0.74]$ \\
Gemma 1.1$\rightarrow$2 & MMLU-Pro & Output JSD & $1.00\,[0.99,1.00]$ & $0.67\,[0.65,0.70]$ \\
Qwen2.5$\rightarrow$3 & MMLU-Pro & Representation drift & --- & --- \\
Llama-3$\rightarrow$3.1 & MMLU-Pro & Representation drift & $0.56\,[0.52,0.60]$ & $0.57\,[0.53,0.60]$ \\
Mistral v0.2$\rightarrow$v0.3 & MMLU-Pro & Representation drift & $0.71\,[0.67,0.75]$ & $0.62\,[0.57,0.65]$ \\
Qwen2$\rightarrow$2.5 & MMLU-Pro & Representation drift & $0.65\,[0.61,0.69]$ & $0.53\,[0.50,0.57]$ \\
Gemma 1.1$\rightarrow$2 & MMLU-Pro & Representation drift & --- & --- \\
\midrule
Qwen2.5$\rightarrow$3 & GPQA & $\mathrm{conf}_{\text{old}}$ & $0.72\,[0.64,0.80]$ & $0.59\,[0.51,0.67]$ \\
Llama-3$\rightarrow$3.1 & GPQA & $\mathrm{conf}_{\text{old}}$ & $0.73\,[0.64,0.82]$ & $0.64\,[0.55,0.72]$ \\
Mistral v0.2$\rightarrow$v0.3 & GPQA & $\mathrm{conf}_{\text{old}}$ & $0.71\,[0.61,0.79]$ & $0.59\,[0.51,0.66]$ \\
Qwen2$\rightarrow$2.5 & GPQA & $\mathrm{conf}_{\text{old}}$ & $0.75\,[0.67,0.83]$ & $0.63\,[0.56,0.70]$ \\
Gemma 1.1$\rightarrow$2 & GPQA & $\mathrm{conf}_{\text{old}}$ & $0.65\,[0.55,0.74]$ & $0.52\,[0.45,0.60]$ \\
Qwen2.5$\rightarrow$3 & GPQA & $\mathrm{conf}_{\text{new}}$ & $0.75\,[0.66,0.82]$ & $0.60\,[0.53,0.67]$ \\
Llama-3$\rightarrow$3.1 & GPQA & $\mathrm{conf}_{\text{new}}$ & $0.66\,[0.56,0.76]$ & $0.60\,[0.51,0.68]$ \\
Mistral v0.2$\rightarrow$v0.3 & GPQA & $\mathrm{conf}_{\text{new}}$ & $0.68\,[0.59,0.78]$ & $0.59\,[0.50,0.67]$ \\
Qwen2$\rightarrow$2.5 & GPQA & $\mathrm{conf}_{\text{new}}$ & $0.69\,[0.59,0.77]$ & $0.58\,[0.51,0.65]$ \\
Gemma 1.1$\rightarrow$2 & GPQA & $\mathrm{conf}_{\text{new}}$ & $0.66\,[0.55,0.75]$ & $0.54\,[0.47,0.61]$ \\
Qwen2.5$\rightarrow$3 & GPQA & Logit margin (old) & $0.71\,[0.63,0.79]$ & $0.59\,[0.51,0.66]$ \\
Llama-3$\rightarrow$3.1 & GPQA & Logit margin (old) & $0.71\,[0.61,0.80]$ & $0.64\,[0.55,0.73]$ \\
Mistral v0.2$\rightarrow$v0.3 & GPQA & Logit margin (old) & $0.71\,[0.61,0.79]$ & $0.58\,[0.51,0.66]$ \\
Qwen2$\rightarrow$2.5 & GPQA & Logit margin (old) & $0.74\,[0.66,0.82]$ & $0.64\,[0.57,0.71]$ \\
Gemma 1.1$\rightarrow$2 & GPQA & Logit margin (old) & $0.65\,[0.55,0.74]$ & $0.52\,[0.45,0.60]$ \\
Qwen2.5$\rightarrow$3 & GPQA & Logit margin (new) & $0.74\,[0.66,0.81]$ & $0.61\,[0.54,0.68]$ \\
Llama-3$\rightarrow$3.1 & GPQA & Logit margin (new) & $0.67\,[0.57,0.76]$ & $0.62\,[0.53,0.71]$ \\
Mistral v0.2$\rightarrow$v0.3 & GPQA & Logit margin (new) & $0.66\,[0.56,0.76]$ & $0.57\,[0.49,0.66]$ \\
Qwen2$\rightarrow$2.5 & GPQA & Logit margin (new) & $0.69\,[0.59,0.77]$ & $0.58\,[0.50,0.65]$ \\
Gemma 1.1$\rightarrow$2 & GPQA & Logit margin (new) & $0.66\,[0.56,0.75]$ & $0.55\,[0.47,0.62]$ \\
Qwen2.5$\rightarrow$3 & GPQA & Attention entropy (old) & $0.55\,[0.45,0.63]$ & $0.51\,[0.42,0.59]$ \\
Llama-3$\rightarrow$3.1 & GPQA & Attention entropy (old) & $0.56\,[0.45,0.66]$ & $0.51\,[0.41,0.59]$ \\
Mistral v0.2$\rightarrow$v0.3 & GPQA & Attention entropy (old) & $0.58\,[0.47,0.69]$ & $0.52\,[0.42,0.62]$ \\
Qwen2$\rightarrow$2.5 & GPQA & Attention entropy (old) & $0.52\,[0.42,0.62]$ & $0.53\,[0.46,0.60]$ \\
Gemma 1.1$\rightarrow$2 & GPQA & Attention entropy (old) & $0.55\,[0.44,0.65]$ & $0.51\,[0.43,0.59]$ \\
Qwen2.5$\rightarrow$3 & GPQA & Attention entropy (new) & $0.70\,[0.60,0.79]$ & $0.62\,[0.54,0.69]$ \\
Llama-3$\rightarrow$3.1 & GPQA & Attention entropy (new) & $0.58\,[0.47,0.68]$ & $0.56\,[0.48,0.64]$ \\
Mistral v0.2$\rightarrow$v0.3 & GPQA & Attention entropy (new) & $0.55\,[0.44,0.66]$ & $0.54\,[0.44,0.65]$ \\
Qwen2$\rightarrow$2.5 & GPQA & Attention entropy (new) & $0.51\,[0.41,0.61]$ & $0.53\,[0.46,0.61]$ \\
Gemma 1.1$\rightarrow$2 & GPQA & Attention entropy (new) & $0.50\,[0.40,0.61]$ & $0.54\,[0.46,0.62]$ \\
Qwen2.5$\rightarrow$3 & GPQA & Output KL (old$\Vert$new) & $0.95\,[0.91,0.98]$ & $0.69\,[0.63,0.74]$ \\
Llama-3$\rightarrow$3.1 & GPQA & Output KL (old$\Vert$new) & $0.82\,[0.74,0.89]$ & $0.74\,[0.66,0.81]$ \\
Mistral v0.2$\rightarrow$v0.3 & GPQA & Output KL (old$\Vert$new) & $0.98\,[0.95,1.00]$ & $0.83\,[0.78,0.87]$ \\
Qwen2$\rightarrow$2.5 & GPQA & Output KL (old$\Vert$new) & $0.92\,[0.86,0.96]$ & $0.73\,[0.68,0.79]$ \\
Gemma 1.1$\rightarrow$2 & GPQA & Output KL (old$\Vert$new) & $0.98\,[0.96,1.00]$ & $0.69\,[0.63,0.74]$ \\
Qwen2.5$\rightarrow$3 & GPQA & Output KL (new$\Vert$old) & $0.94\,[0.90,0.97]$ & $0.70\,[0.65,0.76]$ \\
Llama-3$\rightarrow$3.1 & GPQA & Output KL (new$\Vert$old) & $0.76\,[0.67,0.84]$ & $0.69\,[0.62,0.76]$ \\
Mistral v0.2$\rightarrow$v0.3 & GPQA & Output KL (new$\Vert$old) & $0.90\,[0.85,0.95]$ & $0.78\,[0.72,0.83]$ \\
Qwen2$\rightarrow$2.5 & GPQA & Output KL (new$\Vert$old) & $0.98\,[0.95,1.00]$ & $0.77\,[0.73,0.81]$ \\
Gemma 1.1$\rightarrow$2 & GPQA & Output KL (new$\Vert$old) & $0.95\,[0.92,0.98]$ & $0.71\,[0.65,0.76]$ \\
Qwen2.5$\rightarrow$3 & GPQA & Output JSD & $0.98\,[0.96,1.00]$ & $0.72\,[0.66,0.77]$ \\
Llama-3$\rightarrow$3.1 & GPQA & Output JSD & $0.80\,[0.72,0.88]$ & $0.72\,[0.65,0.80]$ \\
Mistral v0.2$\rightarrow$v0.3 & GPQA & Output JSD & $0.99\,[0.98,1.00]$ & $0.84\,[0.80,0.88]$ \\
Qwen2$\rightarrow$2.5 & GPQA & Output JSD & $0.99\,[0.98,1.00]$ & $0.77\,[0.73,0.82]$ \\
Gemma 1.1$\rightarrow$2 & GPQA & Output JSD & $0.99\,[0.98,1.00]$ & $0.71\,[0.66,0.76]$ \\
Qwen2.5$\rightarrow$3 & GPQA & Representation drift & --- & --- \\
Llama-3$\rightarrow$3.1 & GPQA & Representation drift & $0.52\,[0.41,0.61]$ & $0.53\,[0.45,0.62]$ \\
Mistral v0.2$\rightarrow$v0.3 & GPQA & Representation drift & $0.61\,[0.50,0.72]$ & $0.58\,[0.48,0.68]$ \\
Qwen2$\rightarrow$2.5 & GPQA & Representation drift & $0.59\,[0.50,0.69]$ & $0.58\,[0.51,0.65]$ \\
Gemma 1.1$\rightarrow$2 & GPQA & Representation drift & --- & --- \\
\midrule
Qwen2.5$\rightarrow$3 & GSM8K & $\mathrm{conf}_{\text{old}}$ & $0.68\,[0.58,0.78]$ & $0.66\,[0.56,0.76]$ \\
Llama-3$\rightarrow$3.1 & GSM8K & $\mathrm{conf}_{\text{old}}$ & $0.64\,[0.58,0.70]$ & $0.58\,[0.52,0.64]$ \\
Mistral v0.2$\rightarrow$v0.3 & GSM8K & $\mathrm{conf}_{\text{old}}$ & $0.57\,[0.52,0.63]$ & $0.56\,[0.51,0.61]$ \\
Qwen2$\rightarrow$2.5 & GSM8K & $\mathrm{conf}_{\text{old}}$ & $0.59\,[0.50,0.67]$ & $0.55\,[0.47,0.64]$ \\
Gemma 1.1$\rightarrow$2 & GSM8K & $\mathrm{conf}_{\text{old}}$ & $0.54\,[0.43,0.65]$ & $0.52\,[0.42,0.63]$ \\
Qwen2.5$\rightarrow$3 & GSM8K & $\mathrm{conf}_{\text{new}}$ & $0.76\,[0.66,0.85]$ & $0.74\,[0.64,0.84]$ \\
Llama-3$\rightarrow$3.1 & GSM8K & $\mathrm{conf}_{\text{new}}$ & $0.65\,[0.59,0.71]$ & $0.61\,[0.55,0.66]$ \\
Mistral v0.2$\rightarrow$v0.3 & GSM8K & $\mathrm{conf}_{\text{new}}$ & $0.70\,[0.65,0.74]$ & $0.62\,[0.57,0.66]$ \\
Qwen2$\rightarrow$2.5 & GSM8K & $\mathrm{conf}_{\text{new}}$ & $0.78\,[0.71,0.85]$ & $0.76\,[0.68,0.83]$ \\
Gemma 1.1$\rightarrow$2 & GSM8K & $\mathrm{conf}_{\text{new}}$ & $0.62\,[0.52,0.72]$ & $0.57\,[0.46,0.67]$ \\
Qwen2.5$\rightarrow$3 & GSM8K & Likelihood drift (old traj.) & $0.56\,[0.47,0.66]$ & $0.56\,[0.46,0.65]$ \\
Llama-3$\rightarrow$3.1 & GSM8K & Likelihood drift (old traj.) & $0.58\,[0.52,0.64]$ & $0.54\,[0.48,0.60]$ \\
Mistral v0.2$\rightarrow$v0.3 & GSM8K & Likelihood drift (old traj.) & $0.53\,[0.48,0.59]$ & $0.53\,[0.48,0.58]$ \\
Qwen2$\rightarrow$2.5 & GSM8K & Likelihood drift (old traj.) & $0.51\,[0.43,0.60]$ & $0.53\,[0.44,0.61]$ \\
Gemma 1.1$\rightarrow$2 & GSM8K & Likelihood drift (old traj.) & $0.56\,[0.45,0.66]$ & $0.52\,[0.42,0.61]$ \\
Qwen2.5$\rightarrow$3 & GSM8K & Likelihood drift (new traj.) & $0.63\,[0.52,0.74]$ & $0.63\,[0.51,0.73]$ \\
Llama-3$\rightarrow$3.1 & GSM8K & Likelihood drift (new traj.) & $0.51\,[0.45,0.57]$ & $0.51\,[0.45,0.57]$ \\
Mistral v0.2$\rightarrow$v0.3 & GSM8K & Likelihood drift (new traj.) & $0.52\,[0.46,0.57]$ & $0.51\,[0.46,0.56]$ \\
Qwen2$\rightarrow$2.5 & GSM8K & Likelihood drift (new traj.) & $0.65\,[0.56,0.73]$ & $0.64\,[0.56,0.72]$ \\
Gemma 1.1$\rightarrow$2 & GSM8K & Likelihood drift (new traj.) & $0.52\,[0.41,0.63]$ & $0.53\,[0.43,0.63]$ \\
Qwen2.5$\rightarrow$3 & GSM8K & Top-$k$ token KL (old traj.) & $0.55\,[0.46,0.64]$ & $0.54\,[0.46,0.63]$ \\
Llama-3$\rightarrow$3.1 & GSM8K & Top-$k$ token KL (old traj.) & $0.58\,[0.52,0.64]$ & $0.54\,[0.48,0.60]$ \\
Mistral v0.2$\rightarrow$v0.3 & GSM8K & Top-$k$ token KL (old traj.) & $0.53\,[0.48,0.59]$ & $0.53\,[0.48,0.58]$ \\
Qwen2$\rightarrow$2.5 & GSM8K & Top-$k$ token KL (old traj.) & $0.52\,[0.44,0.61]$ & $0.53\,[0.45,0.61]$ \\
Gemma 1.1$\rightarrow$2 & GSM8K & Top-$k$ token KL (old traj.) & $0.56\,[0.46,0.66]$ & $0.51\,[0.42,0.61]$ \\
Qwen2.5$\rightarrow$3 & GSM8K & Top-$k$ token KL (new traj.) & $0.60\,[0.49,0.71]$ & $0.60\,[0.48,0.72]$ \\
Llama-3$\rightarrow$3.1 & GSM8K & Top-$k$ token KL (new traj.) & $0.60\,[0.53,0.67]$ & $0.58\,[0.52,0.65]$ \\
Mistral v0.2$\rightarrow$v0.3 & GSM8K & Top-$k$ token KL (new traj.) & $0.52\,[0.47,0.58]$ & $0.51\,[0.46,0.56]$ \\
Qwen2$\rightarrow$2.5 & GSM8K & Top-$k$ token KL (new traj.) & $0.66\,[0.57,0.74]$ & $0.65\,[0.56,0.73]$ \\
Gemma 1.1$\rightarrow$2 & GSM8K & Top-$k$ token KL (new traj.) & $0.52\,[0.41,0.63]$ & $0.53\,[0.43,0.63]$ \\
Qwen2.5$\rightarrow$3 & GSM8K & Rep.\ drift, last tok (old traj.) & --- & --- \\
Llama-3$\rightarrow$3.1 & GSM8K & Rep.\ drift, last tok (old traj.) & $0.53\,[0.47,0.59]$ & $0.54\,[0.47,0.60]$ \\
Mistral v0.2$\rightarrow$v0.3 & GSM8K & Rep.\ drift, last tok (old traj.) & $0.54\,[0.49,0.60]$ & $0.52\,[0.46,0.57]$ \\
Qwen2$\rightarrow$2.5 & GSM8K & Rep.\ drift, last tok (old traj.) & $0.50\,[0.42,0.58]$ & $0.51\,[0.43,0.59]$ \\
Gemma 1.1$\rightarrow$2 & GSM8K & Rep.\ drift, last tok (old traj.) & --- & --- \\
Qwen2.5$\rightarrow$3 & GSM8K & Rep.\ drift, last tok (new traj.) & --- & --- \\
Llama-3$\rightarrow$3.1 & GSM8K & Rep.\ drift, last tok (new traj.) & $0.56\,[0.49,0.62]$ & $0.53\,[0.47,0.60]$ \\
Mistral v0.2$\rightarrow$v0.3 & GSM8K & Rep.\ drift, last tok (new traj.) & $0.51\,[0.45,0.56]$ & $0.50\,[0.45,0.55]$ \\
Qwen2$\rightarrow$2.5 & GSM8K & Rep.\ drift, last tok (new traj.) & $0.59\,[0.50,0.69]$ & $0.59\,[0.49,0.67]$ \\
Gemma 1.1$\rightarrow$2 & GSM8K & Rep.\ drift, last tok (new traj.) & --- & --- \\
Qwen2.5$\rightarrow$3 & GSM8K & Rep.\ drift, mean tok (new traj.) & --- & --- \\
Llama-3$\rightarrow$3.1 & GSM8K & Rep.\ drift, mean tok (new traj.) & $0.61\,[0.54,0.67]$ & $0.58\,[0.52,0.64]$ \\
Mistral v0.2$\rightarrow$v0.3 & GSM8K & Rep.\ drift, mean tok (new traj.) & $0.50\,[0.45,0.56]$ & $0.51\,[0.46,0.56]$ \\
Qwen2$\rightarrow$2.5 & GSM8K & Rep.\ drift, mean tok (new traj.) & $0.77\,[0.70,0.83]$ & $0.74\,[0.66,0.80]$ \\
Gemma 1.1$\rightarrow$2 & GSM8K & Rep.\ drift, mean tok (new traj.) & --- & --- \\
\midrule
Qwen2.5$\rightarrow$3 & MATH-full & $\mathrm{conf}_{\text{old}}$ & $0.55\,[0.51,0.59]$ & $0.52\,[0.48,0.57]$ \\
Llama-3$\rightarrow$3.1 & MATH-full & $\mathrm{conf}_{\text{old}}$ & $0.62\,[0.58,0.66]$ & $0.53\,[0.50,0.57]$ \\
Mistral v0.2$\rightarrow$v0.3 & MATH-full & $\mathrm{conf}_{\text{old}}$ & $0.60\,[0.54,0.65]$ & $0.59\,[0.55,0.62]$ \\
Qwen2$\rightarrow$2.5 & MATH-full & $\mathrm{conf}_{\text{old}}$ & $0.56\,[0.50,0.61]$ & $0.53\,[0.48,0.58]$ \\
Gemma 1.1$\rightarrow$2 & MATH-full & $\mathrm{conf}_{\text{old}}$ & $0.70\,[0.64,0.75]$ & $0.55\,[0.50,0.60]$ \\
Qwen2.5$\rightarrow$3 & MATH-full & $\mathrm{conf}_{\text{new}}$ & $0.68\,[0.64,0.72]$ & $0.63\,[0.59,0.67]$ \\
Llama-3$\rightarrow$3.1 & MATH-full & $\mathrm{conf}_{\text{new}}$ & $0.62\,[0.57,0.66]$ & $0.58\,[0.54,0.62]$ \\
Mistral v0.2$\rightarrow$v0.3 & MATH-full & $\mathrm{conf}_{\text{new}}$ & $0.63\,[0.58,0.68]$ & $0.51\,[0.47,0.55]$ \\
Qwen2$\rightarrow$2.5 & MATH-full & $\mathrm{conf}_{\text{new}}$ & $0.57\,[0.52,0.62]$ & $0.54\,[0.49,0.58]$ \\
Gemma 1.1$\rightarrow$2 & MATH-full & $\mathrm{conf}_{\text{new}}$ & $0.67\,[0.61,0.73]$ & $0.55\,[0.50,0.60]$ \\
Qwen2.5$\rightarrow$3 & MATH-full & Likelihood drift (old traj.) & $0.57\,[0.53,0.61]$ & $0.56\,[0.51,0.60]$ \\
Llama-3$\rightarrow$3.1 & MATH-full & Likelihood drift (old traj.) & $0.51\,[0.46,0.55]$ & $0.52\,[0.48,0.55]$ \\
Mistral v0.2$\rightarrow$v0.3 & MATH-full & Likelihood drift (old traj.) & $0.58\,[0.52,0.63]$ & $0.64\,[0.60,0.67]$ \\
Qwen2$\rightarrow$2.5 & MATH-full & Likelihood drift (old traj.) & $0.50\,[0.45,0.55]$ & $0.53\,[0.48,0.58]$ \\
Gemma 1.1$\rightarrow$2 & MATH-full & Likelihood drift (old traj.) & $0.53\,[0.48,0.59]$ & $0.51\,[0.46,0.56]$ \\
Qwen2.5$\rightarrow$3 & MATH-full & Likelihood drift (new traj.) & $0.62\,[0.57,0.66]$ & $0.59\,[0.54,0.63]$ \\
Llama-3$\rightarrow$3.1 & MATH-full & Likelihood drift (new traj.) & $0.54\,[0.50,0.58]$ & $0.55\,[0.52,0.59]$ \\
Mistral v0.2$\rightarrow$v0.3 & MATH-full & Likelihood drift (new traj.) & $0.60\,[0.55,0.65]$ & $0.59\,[0.55,0.63]$ \\
Qwen2$\rightarrow$2.5 & MATH-full & Likelihood drift (new traj.) & $0.54\,[0.49,0.59]$ & $0.51\,[0.46,0.56]$ \\
Gemma 1.1$\rightarrow$2 & MATH-full & Likelihood drift (new traj.) & $0.56\,[0.50,0.61]$ & $0.57\,[0.51,0.62]$ \\
Qwen2.5$\rightarrow$3 & MATH-full & Top-$k$ token KL (old traj.) & $0.57\,[0.53,0.62]$ & $0.56\,[0.51,0.60]$ \\
Llama-3$\rightarrow$3.1 & MATH-full & Top-$k$ token KL (old traj.) & $0.50\,[0.46,0.54]$ & $0.50\,[0.47,0.54]$ \\
Mistral v0.2$\rightarrow$v0.3 & MATH-full & Top-$k$ token KL (old traj.) & $0.58\,[0.53,0.63]$ & $0.64\,[0.60,0.67]$ \\
Qwen2$\rightarrow$2.5 & MATH-full & Top-$k$ token KL (old traj.) & $0.51\,[0.46,0.56]$ & $0.52\,[0.47,0.57]$ \\
Gemma 1.1$\rightarrow$2 & MATH-full & Top-$k$ token KL (old traj.) & $0.54\,[0.48,0.60]$ & $0.50\,[0.45,0.56]$ \\
Qwen2.5$\rightarrow$3 & MATH-full & Top-$k$ token KL (new traj.) & $0.61\,[0.57,0.66]$ & $0.58\,[0.54,0.62]$ \\
Llama-3$\rightarrow$3.1 & MATH-full & Top-$k$ token KL (new traj.) & $0.58\,[0.53,0.62]$ & $0.57\,[0.53,0.61]$ \\
Mistral v0.2$\rightarrow$v0.3 & MATH-full & Top-$k$ token KL (new traj.) & $0.60\,[0.55,0.65]$ & $0.59\,[0.56,0.63]$ \\
Qwen2$\rightarrow$2.5 & MATH-full & Top-$k$ token KL (new traj.) & $0.54\,[0.49,0.59]$ & $0.51\,[0.46,0.56]$ \\
Gemma 1.1$\rightarrow$2 & MATH-full & Top-$k$ token KL (new traj.) & $0.55\,[0.50,0.61]$ & $0.56\,[0.50,0.61]$ \\
Qwen2.5$\rightarrow$3 & MATH-full & Rep.\ drift, last tok (old traj.) & --- & --- \\
Llama-3$\rightarrow$3.1 & MATH-full & Rep.\ drift, last tok (old traj.) & $0.58\,[0.54,0.62]$ & $0.51\,[0.48,0.55]$ \\
Mistral v0.2$\rightarrow$v0.3 & MATH-full & Rep.\ drift, last tok (old traj.) & $0.51\,[0.45,0.56]$ & $0.54\,[0.51,0.58]$ \\
Qwen2$\rightarrow$2.5 & MATH-full & Rep.\ drift, last tok (old traj.) & $0.50\,[0.45,0.55]$ & $0.54\,[0.49,0.59]$ \\
Gemma 1.1$\rightarrow$2 & MATH-full & Rep.\ drift, last tok (old traj.) & --- & --- \\
Qwen2.5$\rightarrow$3 & MATH-full & Rep.\ drift, last tok (new traj.) & --- & --- \\
Llama-3$\rightarrow$3.1 & MATH-full & Rep.\ drift, last tok (new traj.) & $0.59\,[0.54,0.63]$ & $0.52\,[0.48,0.57]$ \\
Mistral v0.2$\rightarrow$v0.3 & MATH-full & Rep.\ drift, last tok (new traj.) & $0.51\,[0.46,0.56]$ & $0.53\,[0.50,0.57]$ \\
Qwen2$\rightarrow$2.5 & MATH-full & Rep.\ drift, last tok (new traj.) & $0.56\,[0.51,0.61]$ & $0.51\,[0.46,0.56]$ \\
Gemma 1.1$\rightarrow$2 & MATH-full & Rep.\ drift, last tok (new traj.) & --- & --- \\
Qwen2.5$\rightarrow$3 & MATH-full & Rep.\ drift, mean tok (new traj.) & --- & --- \\
Llama-3$\rightarrow$3.1 & MATH-full & Rep.\ drift, mean tok (new traj.) & $0.60\,[0.56,0.63]$ & $0.53\,[0.49,0.57]$ \\
Mistral v0.2$\rightarrow$v0.3 & MATH-full & Rep.\ drift, mean tok (new traj.) & $0.58\,[0.53,0.63]$ & $0.53\,[0.50,0.57]$ \\
Qwen2$\rightarrow$2.5 & MATH-full & Rep.\ drift, mean tok (new traj.) & $0.67\,[0.62,0.71]$ & $0.57\,[0.53,0.62]$ \\
Gemma 1.1$\rightarrow$2 & MATH-full & Rep.\ drift, mean tok (new traj.) & --- & --- \\
\midrule
Qwen2.5$\rightarrow$3 & HumanEval & $\mathrm{conf}_{\text{old}}$ & $0.55\,[0.41,0.69]$ & $0.61\,[0.48,0.74]$ \\
Llama-3$\rightarrow$3.1 & HumanEval & $\mathrm{conf}_{\text{old}}$ & $0.55\,[0.38,0.71]$ & $0.57\,[0.41,0.71]$ \\
Mistral v0.2$\rightarrow$v0.3 & HumanEval & $\mathrm{conf}_{\text{old}}$ & $0.56\,[0.35,0.79]$ & $0.62\,[0.42,0.80]$ \\
Qwen2$\rightarrow$2.5 & HumanEval & $\mathrm{conf}_{\text{old}}$ & $0.64\,[0.44,0.83]$ & $0.58\,[0.37,0.77]$ \\
Gemma 1.1$\rightarrow$2 & HumanEval & $\mathrm{conf}_{\text{old}}$ & $0.55\,[0.41,0.69]$ & $0.65\,[0.52,0.76]$ \\
Qwen2.5$\rightarrow$3 & HumanEval & $\mathrm{conf}_{\text{new}}$ & $0.81\,[0.67,0.93]$ & $0.79\,[0.65,0.91]$ \\
Llama-3$\rightarrow$3.1 & HumanEval & $\mathrm{conf}_{\text{new}}$ & $0.75\,[0.60,0.88]$ & $0.68\,[0.53,0.82]$ \\
Mistral v0.2$\rightarrow$v0.3 & HumanEval & $\mathrm{conf}_{\text{new}}$ & $0.67\,[0.51,0.83]$ & $0.69\,[0.56,0.81]$ \\
Qwen2$\rightarrow$2.5 & HumanEval & $\mathrm{conf}_{\text{new}}$ & $0.75\,[0.58,0.90]$ & $0.69\,[0.51,0.85]$ \\
Gemma 1.1$\rightarrow$2 & HumanEval & $\mathrm{conf}_{\text{new}}$ & $0.61\,[0.45,0.76]$ & $0.55\,[0.42,0.68]$ \\
Qwen2.5$\rightarrow$3 & HumanEval & Likelihood drift (old traj.) & $0.54\,[0.36,0.70]$ & $0.54\,[0.37,0.71]$ \\
Llama-3$\rightarrow$3.1 & HumanEval & Likelihood drift (old traj.) & $0.56\,[0.41,0.68]$ & $0.60\,[0.47,0.72]$ \\
Mistral v0.2$\rightarrow$v0.3 & HumanEval & Likelihood drift (old traj.) & $0.50\,[0.30,0.70]$ & $0.50\,[0.31,0.69]$ \\
Qwen2$\rightarrow$2.5 & HumanEval & Likelihood drift (old traj.) & $0.58\,[0.40,0.76]$ & $0.52\,[0.33,0.71]$ \\
Gemma 1.1$\rightarrow$2 & HumanEval & Likelihood drift (old traj.) & $0.69\,[0.55,0.81]$ & $0.72\,[0.62,0.82]$ \\
Qwen2.5$\rightarrow$3 & HumanEval & Likelihood drift (new traj.) & $0.59\,[0.42,0.74]$ & $0.59\,[0.42,0.74]$ \\
Llama-3$\rightarrow$3.1 & HumanEval & Likelihood drift (new traj.) & $0.73\,[0.61,0.84]$ & $0.67\,[0.54,0.78]$ \\
Mistral v0.2$\rightarrow$v0.3 & HumanEval & Likelihood drift (new traj.) & $0.54\,[0.34,0.76]$ & $0.57\,[0.38,0.77]$ \\
Qwen2$\rightarrow$2.5 & HumanEval & Likelihood drift (new traj.) & $0.80\,[0.65,0.92]$ & $0.76\,[0.62,0.88]$ \\
Gemma 1.1$\rightarrow$2 & HumanEval & Likelihood drift (new traj.) & $0.64\,[0.49,0.78]$ & $0.52\,[0.39,0.65]$ \\
Qwen2.5$\rightarrow$3 & HumanEval & Top-$k$ token KL (old traj.) & $0.51\,[0.33,0.68]$ & $0.52\,[0.34,0.69]$ \\
Llama-3$\rightarrow$3.1 & HumanEval & Top-$k$ token KL (old traj.) & $0.63\,[0.48,0.77]$ & $0.64\,[0.50,0.77]$ \\
Mistral v0.2$\rightarrow$v0.3 & HumanEval & Top-$k$ token KL (old traj.) & $0.50\,[0.30,0.70]$ & $0.51\,[0.31,0.69]$ \\
Qwen2$\rightarrow$2.5 & HumanEval & Top-$k$ token KL (old traj.) & $0.61\,[0.41,0.78]$ & $0.54\,[0.35,0.71]$ \\
Gemma 1.1$\rightarrow$2 & HumanEval & Top-$k$ token KL (old traj.) & $0.67\,[0.53,0.79]$ & $0.71\,[0.60,0.81]$ \\
Qwen2.5$\rightarrow$3 & HumanEval & Top-$k$ token KL (new traj.) & $0.58\,[0.42,0.73]$ & $0.60\,[0.44,0.74]$ \\
Llama-3$\rightarrow$3.1 & HumanEval & Top-$k$ token KL (new traj.) & $0.70\,[0.57,0.83]$ & $0.66\,[0.53,0.77]$ \\
Mistral v0.2$\rightarrow$v0.3 & HumanEval & Top-$k$ token KL (new traj.) & $0.52\,[0.32,0.75]$ & $0.56\,[0.37,0.75]$ \\
Qwen2$\rightarrow$2.5 & HumanEval & Top-$k$ token KL (new traj.) & $0.79\,[0.64,0.91]$ & $0.75\,[0.60,0.87]$ \\
Gemma 1.1$\rightarrow$2 & HumanEval & Top-$k$ token KL (new traj.) & $0.64\,[0.49,0.78]$ & $0.51\,[0.39,0.64]$ \\
Qwen2.5$\rightarrow$3 & HumanEval & Rep.\ drift, last tok (old traj.) & --- & --- \\
Llama-3$\rightarrow$3.1 & HumanEval & Rep.\ drift, last tok (old traj.) & $0.50\,[0.35,0.65]$ & $0.50\,[0.37,0.65]$ \\
Mistral v0.2$\rightarrow$v0.3 & HumanEval & Rep.\ drift, last tok (old traj.) & $0.55\,[0.34,0.75]$ & $0.57\,[0.37,0.76]$ \\
Qwen2$\rightarrow$2.5 & HumanEval & Rep.\ drift, last tok (old traj.) & $0.62\,[0.45,0.78]$ & $0.61\,[0.45,0.77]$ \\
Gemma 1.1$\rightarrow$2 & HumanEval & Rep.\ drift, last tok (old traj.) & --- & --- \\
Qwen2.5$\rightarrow$3 & HumanEval & Rep.\ drift, last tok (new traj.) & --- & --- \\
Llama-3$\rightarrow$3.1 & HumanEval & Rep.\ drift, last tok (new traj.) & $0.57\,[0.41,0.73]$ & $0.59\,[0.43,0.73]$ \\
Mistral v0.2$\rightarrow$v0.3 & HumanEval & Rep.\ drift, last tok (new traj.) & $0.52\,[0.28,0.75]$ & $0.53\,[0.32,0.73]$ \\
Qwen2$\rightarrow$2.5 & HumanEval & Rep.\ drift, last tok (new traj.) & $0.50\,[0.32,0.68]$ & $0.50\,[0.32,0.67]$ \\
Gemma 1.1$\rightarrow$2 & HumanEval & Rep.\ drift, last tok (new traj.) & --- & --- \\
Qwen2.5$\rightarrow$3 & HumanEval & Rep.\ drift, mean tok (new traj.) & --- & --- \\
Llama-3$\rightarrow$3.1 & HumanEval & Rep.\ drift, mean tok (new traj.) & $0.54\,[0.39,0.69]$ & $0.55\,[0.40,0.69]$ \\
Mistral v0.2$\rightarrow$v0.3 & HumanEval & Rep.\ drift, mean tok (new traj.) & $0.65\,[0.45,0.83]$ & $0.65\,[0.47,0.81]$ \\
Qwen2$\rightarrow$2.5 & HumanEval & Rep.\ drift, mean tok (new traj.) & $0.60\,[0.42,0.75]$ & $0.62\,[0.46,0.76]$ \\
Gemma 1.1$\rightarrow$2 & HumanEval & Rep.\ drift, mean tok (new traj.) & --- & --- \\
\midrule
Qwen2.5$\rightarrow$3 & MBPP & $\mathrm{conf}_{\text{old}}$ & $0.67\,[0.57,0.76]$ & $0.58\,[0.48,0.68]$ \\
Llama-3$\rightarrow$3.1 & MBPP & $\mathrm{conf}_{\text{old}}$ & $0.70\,[0.59,0.80]$ & $0.54\,[0.45,0.64]$ \\
Mistral v0.2$\rightarrow$v0.3 & MBPP & $\mathrm{conf}_{\text{old}}$ & $0.59\,[0.47,0.70]$ & $0.67\,[0.59,0.75]$ \\
Qwen2$\rightarrow$2.5 & MBPP & $\mathrm{conf}_{\text{old}}$ & $0.69\,[0.61,0.77]$ & $0.55\,[0.47,0.63]$ \\
Gemma 1.1$\rightarrow$2 & MBPP & $\mathrm{conf}_{\text{old}}$ & $0.52\,[0.40,0.65]$ & $0.61\,[0.51,0.71]$ \\
Qwen2.5$\rightarrow$3 & MBPP & $\mathrm{conf}_{\text{new}}$ & $0.68\,[0.59,0.76]$ & $0.61\,[0.53,0.69]$ \\
Llama-3$\rightarrow$3.1 & MBPP & $\mathrm{conf}_{\text{new}}$ & $0.76\,[0.65,0.85]$ & $0.61\,[0.51,0.71]$ \\
Mistral v0.2$\rightarrow$v0.3 & MBPP & $\mathrm{conf}_{\text{new}}$ & $0.81\,[0.71,0.89]$ & $0.60\,[0.48,0.70]$ \\
Qwen2$\rightarrow$2.5 & MBPP & $\mathrm{conf}_{\text{new}}$ & $0.67\,[0.56,0.77]$ & $0.57\,[0.46,0.67]$ \\
Gemma 1.1$\rightarrow$2 & MBPP & $\mathrm{conf}_{\text{new}}$ & $0.62\,[0.48,0.74]$ & $0.59\,[0.47,0.71]$ \\
Qwen2.5$\rightarrow$3 & MBPP & Likelihood drift (old traj.) & $0.67\,[0.58,0.75]$ & $0.66\,[0.58,0.74]$ \\
Llama-3$\rightarrow$3.1 & MBPP & Likelihood drift (old traj.) & $0.52\,[0.40,0.63]$ & $0.55\,[0.46,0.65]$ \\
Mistral v0.2$\rightarrow$v0.3 & MBPP & Likelihood drift (old traj.) & $0.56\,[0.40,0.71]$ & $0.53\,[0.39,0.68]$ \\
Qwen2$\rightarrow$2.5 & MBPP & Likelihood drift (old traj.) & $0.78\,[0.71,0.85]$ & $0.66\,[0.57,0.73]$ \\
Gemma 1.1$\rightarrow$2 & MBPP & Likelihood drift (old traj.) & $0.57\,[0.44,0.72]$ & $0.50\,[0.38,0.62]$ \\
Qwen2.5$\rightarrow$3 & MBPP & Likelihood drift (new traj.) & $0.75\,[0.67,0.82]$ & $0.66\,[0.59,0.74]$ \\
Llama-3$\rightarrow$3.1 & MBPP & Likelihood drift (new traj.) & $0.74\,[0.64,0.84]$ & $0.72\,[0.61,0.82]$ \\
Mistral v0.2$\rightarrow$v0.3 & MBPP & Likelihood drift (new traj.) & $0.63\,[0.48,0.77]$ & $0.55\,[0.41,0.69]$ \\
Qwen2$\rightarrow$2.5 & MBPP & Likelihood drift (new traj.) & $0.75\,[0.67,0.84]$ & $0.63\,[0.54,0.73]$ \\
Gemma 1.1$\rightarrow$2 & MBPP & Likelihood drift (new traj.) & $0.59\,[0.47,0.70]$ & $0.51\,[0.40,0.62]$ \\
Qwen2.5$\rightarrow$3 & MBPP & Top-$k$ token KL (old traj.) & $0.68\,[0.58,0.76]$ & $0.67\,[0.58,0.75]$ \\
Llama-3$\rightarrow$3.1 & MBPP & Top-$k$ token KL (old traj.) & $0.51\,[0.41,0.62]$ & $0.53\,[0.43,0.62]$ \\
Mistral v0.2$\rightarrow$v0.3 & MBPP & Top-$k$ token KL (old traj.) & $0.56\,[0.40,0.72]$ & $0.53\,[0.39,0.68]$ \\
Qwen2$\rightarrow$2.5 & MBPP & Top-$k$ token KL (old traj.) & $0.64\,[0.54,0.73]$ & $0.54\,[0.44,0.64]$ \\
Gemma 1.1$\rightarrow$2 & MBPP & Top-$k$ token KL (old traj.) & $0.55\,[0.42,0.70]$ & $0.52\,[0.40,0.63]$ \\
Qwen2.5$\rightarrow$3 & MBPP & Top-$k$ token KL (new traj.) & $0.74\,[0.66,0.81]$ & $0.66\,[0.58,0.74]$ \\
Llama-3$\rightarrow$3.1 & MBPP & Top-$k$ token KL (new traj.) & $0.72\,[0.62,0.81]$ & $0.68\,[0.57,0.78]$ \\
Mistral v0.2$\rightarrow$v0.3 & MBPP & Top-$k$ token KL (new traj.) & $0.64\,[0.49,0.77]$ & $0.55\,[0.40,0.69]$ \\
Qwen2$\rightarrow$2.5 & MBPP & Top-$k$ token KL (new traj.) & $0.79\,[0.70,0.87]$ & $0.67\,[0.58,0.76]$ \\
Gemma 1.1$\rightarrow$2 & MBPP & Top-$k$ token KL (new traj.) & $0.59\,[0.47,0.70]$ & $0.51\,[0.40,0.62]$ \\
Qwen2.5$\rightarrow$3 & MBPP & Rep.\ drift, last tok (old traj.) & --- & --- \\
Llama-3$\rightarrow$3.1 & MBPP & Rep.\ drift, last tok (old traj.) & $0.55\,[0.44,0.66]$ & $0.52\,[0.42,0.62]$ \\
Mistral v0.2$\rightarrow$v0.3 & MBPP & Rep.\ drift, last tok (old traj.) & $0.58\,[0.43,0.73]$ & $0.56\,[0.42,0.69]$ \\
Qwen2$\rightarrow$2.5 & MBPP & Rep.\ drift, last tok (old traj.) & $0.67\,[0.57,0.76]$ & $0.59\,[0.49,0.68]$ \\
Gemma 1.1$\rightarrow$2 & MBPP & Rep.\ drift, last tok (old traj.) & --- & --- \\
Qwen2.5$\rightarrow$3 & MBPP & Rep.\ drift, last tok (new traj.) & --- & --- \\
Llama-3$\rightarrow$3.1 & MBPP & Rep.\ drift, last tok (new traj.) & $0.56\,[0.44,0.68]$ & $0.52\,[0.41,0.63]$ \\
Mistral v0.2$\rightarrow$v0.3 & MBPP & Rep.\ drift, last tok (new traj.) & $0.71\,[0.56,0.84]$ & $0.64\,[0.50,0.77]$ \\
Qwen2$\rightarrow$2.5 & MBPP & Rep.\ drift, last tok (new traj.) & $0.53\,[0.39,0.65]$ & $0.52\,[0.40,0.66]$ \\
Gemma 1.1$\rightarrow$2 & MBPP & Rep.\ drift, last tok (new traj.) & --- & --- \\
Qwen2.5$\rightarrow$3 & MBPP & Rep.\ drift, mean tok (new traj.) & --- & --- \\
Llama-3$\rightarrow$3.1 & MBPP & Rep.\ drift, mean tok (new traj.) & $0.58\,[0.47,0.69]$ & $0.52\,[0.41,0.63]$ \\
Mistral v0.2$\rightarrow$v0.3 & MBPP & Rep.\ drift, mean tok (new traj.) & $0.60\,[0.43,0.76]$ & $0.51\,[0.37,0.64]$ \\
Qwen2$\rightarrow$2.5 & MBPP & Rep.\ drift, mean tok (new traj.) & $0.57\,[0.45,0.68]$ & $0.53\,[0.42,0.64]$ \\
Gemma 1.1$\rightarrow$2 & MBPP & Rep.\ drift, mean tok (new traj.) & --- & --- \\
\end{longtable}
}

\subsection{A Negative-Flip Case Study}

To show concretely what a negative flip looks like, we reproduce the full reasoning for the GSM8K example in Figure~\ref{fig:flip-examples} (Qwen2.5$\rightarrow$3). The outputs are lightly reformatted from the models' Markdown; the numbers and reasoning steps are unchanged.

\paragraph{Question.} Judy teaches 5 dance classes, every day, on the weekdays and 8 classes on Saturday. If each class has 15 students and she charges \$15.00 per student, how much money does she make in 1 week?

\paragraph{Old model ($M_{\text{old}}$, Qwen2.5) --- correct.} 5 classes per weekday $\times$ 5 weekdays $= 25$ classes, plus 8 on Saturday $= 33$ classes. With 15 students per class, $33 \times 15 = 495$ students. At \$15 per student, $495 \times \$15 = \$7425$. \textbf{Final answer: 7425} ($\checkmark$).

\paragraph{New model ($M_{\text{new}}$, Qwen3) --- incorrect.} ``Weekdays (Monday to Friday): 5 days $\times$ 1 class per day $= 5$ classes,'' plus 8 on Saturday $= 13$ classes. With 15 students per class, $13 \times 15 = 195$ students. At \$15 per student, $195 \times \$15 = \$2925$. \textbf{Final answer: 2925} ($\times$).

\paragraph{What regressed.} The new model misreads ``5 dance classes every day on the weekdays'' as five weekdays with a single class each ($5$ classes total) rather than five classes on each of the five weekdays ($25$). The regression is therefore a reading-comprehension error rather than a calculation error.

%% file: references.bib
@article{cacioli2026beyond,
  title={Beyond the Mean: Within-Model Reliable Change Detection for LLM Evaluation},
  author={Cacioli, Jon-Paul},
  journal={arXiv preprint arXiv:2604.27405},
  year={2026},
  url = {https://arxiv.org/abs/2604.27405},
}

@inproceedings{echterhoff-etal-2024-muscle,
    title = "{MUSCLE}: A Model Update Strategy for Compatible {LLM} Evolution",
    author = "Echterhoff, Jessica Maria  and
      Faghri, Fartash  and
      Vemulapalli, Raviteja  and
      Hu, Ting-Yao  and
      Li, Chun-Liang  and
      Tuzel, Oncel  and
      Pouransari, Hadi",
    editor = "Al-Onaizan, Yaser  and
      Bansal, Mohit  and
      Chen, Yun-Nung",
    booktitle = "Findings of the Association for Computational Linguistics: EMNLP 2024",
    month = nov,
    year = "2024",
    address = "Miami, Florida, USA",
    publisher = "Association for Computational Linguistics",
    url = "https://aclanthology.org/2024.findings-emnlp.430/",
    doi = "10.18653/v1/2024.findings-emnlp.430",
    pages = "7320--7332",
}

@article{shenfeld2025rlrazor,
  title={RL's Razor: Why Online Reinforcement Learning Forgets Less}, 
  author={Idan Shenfeld and Jyothish Pari and Pulkit Agrawal},
  journal={arXiv preprint arXiv:2509.04259},
  year={2025},
  url={https://arxiv.org/abs/2509.04259}, 
}

@article{chen2025retaining,
  title={Retaining by doing: The role of on-policy data in mitigating forgetting},
  author={Chen, Howard and Razin, Noam and Narasimhan, Karthik and Chen, Danqi},
  journal={arXiv preprint arXiv:2510.18874},
  year={2025},
  url={https://arxiv.org/abs/2510.18874}, 
}

@article{jin2026demystifying,
  title={Demystifying language model forgetting with low-rank example associations},
  author={Jin, Xisen and Ren, Xiang},
  journal={Advances in Neural Information Processing Systems},
  volume={38},
  pages={4313--4349},
  year={2026},
  url={https://arxiv.org/abs/2406.14026}, 
}

@article{jin2024forget,
  title={What will my model forget? forecasting forgotten examples in language model refinement},
  author={Jin, Xisen and Ren, Xiang},
  journal={arXiv preprint arXiv:2402.01865},
  year={2024},
  url={https://arxiv.org/abs/2402.01865}, 
}

@inproceedings{xie-etal-2021-regression,
    title = {Regression Bugs Are In Your Model! Measuring, Reducing and Analyzing Regressions In NLP Model Updates},
    author = {Xie, Yuqing  and
      Lai, Yi-An  and
      Xiong, Yuanjun  and
      Zhang, Yi  and
      Soatto, Stefano},
    booktitle = {Proceedings of the 59th Annual Meeting of the Association for Computational Linguistics and the 11th International Joint Conference on Natural Language Processing (Volume 1: Long Papers)},
    month = {August},
    year = {2021},
    publisher = {Association for Computational Linguistics},
    url = {https://aclanthology.org/2021.acl-long.515/},
    doi = {10.18653/v1/2021.acl-long.515},
    pages = {6589--6602},
}

@inproceedings{Yan_2021_CVPR,
    author    = {Yan, Sijie and Xiong, Yuanjun and Kundu, Kaustav and Yang, Shuo and Deng, Siqi and Wang, Meng and Xia, Wei and Soatto, Stefano},
    title     = {Positive-Congruent Training: Towards Regression-Free Model Updates},
    booktitle = {Proceedings of the IEEE/CVF Conference on Computer Vision and Pattern Recognition (CVPR)},
    month     = {June},
    year      = {2021},
    pages     = {14299-14308},
    url       = {https://arxiv.org/abs/2011.09161},
}

@article{belrose2023tunedlens,
  title={Eliciting latent predictions from transformers with the tuned lens},
  author={Belrose, Nora and Ostrovsky, Igor and McKinney, Lev and Furman, Zach and Smith, Logan and Halawi, Danny and Biderman, Stella and Steinhardt, Jacob},
  journal={arXiv preprint arXiv:2303.08112},
  year={2023},
  url={https://arxiv.org/abs/2303.08112}, 
}

@article{ostmeier2026headentropy,
  title={Attention Head Entropy of LLMs Predicts Answer Correctness},
  author={Ostmeier, Sophie and Axelrod, Brian and Varma, Maya and Aali, Asad and Zhang, Yabin and Paschali, Magdalini and Koyejo, Sanmi and Langlotz, Curtis and Chaudhari, Akshay},
  journal={arXiv preprint arXiv:2602.13699},
  year={2026},
  url={https://arxiv.org/abs/2602.13699}, 
}

@article{chen2024inside,
  title={INSIDE: LLMs' internal states retain the power of hallucination detection},
  author={Chen, Chao and Liu, Kai and Chen, Ze and Gu, Yi and Wu, Yue and Tao, Mingyuan and Fu, Zhihang and Ye, Jieping},
  journal={arXiv preprint arXiv:2402.03744},
  year={2024},
  url={https://arxiv.org/abs/2402.03744}, 
}

@inproceedings{sriramanan2024llmcheck,
 author = {Sriramanan, Gaurang and Bharti, Siddhant and Sadasivan, Vinu Sankar and Saha, Shoumik and Kattakinda, Priyatham and Feizi, Soheil},
 booktitle = {Advances in Neural Information Processing Systems},
 doi = {10.52202/079017-1077},
 editor = {A. Globerson and L. Mackey and D. Belgrave and A. Fan and U. Paquet and J. Tomczak and C. Zhang},
 pages = {34188--34216},
 publisher = {Curran Associates, Inc.},
 title = {LLM-Check: Investigating Detection of Hallucinations in Large Language Models},
 url = {https://proceedings.neurips.cc/paper_files/paper/2024/file/3c1e1fdf305195cd620c118aaa9717ad-Paper-Conference.pdf},
 volume = {37},
 year = {2024}
}

@inproceedings{swayamdipta2020cartography,
  title={Dataset cartography: Mapping and diagnosing datasets with training dynamics},
  author={Swayamdipta, Swabha and Schwartz, Roy and Lourie, Nicholas and Wang, Yizhong and Hajishirzi, Hannaneh and Smith, Noah A and Choi, Yejin},
  booktitle={Proceedings of the 2020 Conference on Empirical Methods in Natural Language Processing (EMNLP)},
  pages={9275--9293},
  year={2020},
  url = {https://arxiv.org/abs/2009.10795},
}

@inproceedings{ricci2026mpt,
  title={Mitigating Negative Flips via Margin Preserving Training}, 
  author={Simone Ricci and Niccolò Biondi and Federico Pernici and Alberto Del Bimbo},
  booktitle={Proceedings of the AAAI Conference on Artificial Intelligence},
  pages={8721--8730},
  year={2026},
  url = {https://arxiv.org/abs/2511.08322},
}

@inproceedings{cai2023gatedfusion,
  title={Improving prediction backward-compatiblility in nlp model upgrade with gated fusion},
  author={Lai, Yi-An and Mansimov, Elman and Xie, Yuqing and Zhang, Yi},
  booktitle={Findings of the Association for Computational Linguistics: EACL 2023},
  pages={1010--1022},
  year={2023},
  url = {https://arxiv.org/abs/2302.02080},
}

@article{farquhar2024semantic,
  title={Detecting hallucinations in large language models using semantic entropy},
  author={Farquhar, Sebastian and Kossen, Jannik and Kuhn, Lorenz and Gal, Yarin},
  journal={Nature},
  volume={630},
  number={8017},
  pages={625--630},
  year={2024},
  publisher={Nature Publishing Group UK London},
  url = {https://www.nature.com/articles/s41586-024-07421-0},
}

@article{kossen2024seps,
  title={Semantic entropy probes: Robust and cheap hallucination detection in llms},
  author={Kossen, Jannik and Han, Jiatong and Razzak, Muhammed and Schut, Lisa and Malik, Shreshth and Gal, Yarin},
  journal={arXiv preprint arXiv:2406.15927},
  year={2024},
  url = {https://arxiv.org/abs/2406.15927},
}

@inproceedings{azaria2023saplma,
  title={The internal state of an LLM knows when it’s lying},
  author={Azaria, Amos and Mitchell, Tom},
  booktitle={Findings of the Association for Computational Linguistics: EMNLP 2023},
  pages={967--976},
  year={2023},
  url = {https://arxiv.org/abs/2304.13734},
}

@article{phillips2026entropy,
  title={Entropy Alone is Insufficient for Safe Selective Prediction in LLMs},
  author={Phillips, Edward and Gustafsson, Fredrik K and Wu, Sean and Thakur, Anshul and Clifton, David A},
  journal={arXiv preprint arXiv:2603.21172},
  year={2026},
  url = {https://arxiv.org/abs/2603.21172},
}

@misc{hendrycks2021math,
      title={Measuring Mathematical Problem Solving With the MATH Dataset}, 
      author={Dan Hendrycks and Collin Burns and Saurav Kadavath and Akul Arora and Steven Basart and Eric Tang and Dawn Song and Jacob Steinhardt},
      year={2021},
      primaryClass={cs.LG},
      url={https://arxiv.org/abs/2103.03874}, 
}

@misc{chen2021humaneval,
      title={Evaluating Large Language Models Trained on Code}, 
      author={Mark Chen and Jerry Tworek and Heewoo Jun and Qiming Yuan and Henrique Ponde de Oliveira Pinto and Jared Kaplan and Harri Edwards and Yuri Burda and Nicholas Joseph and Greg Brockman and others},
      year={2021},
      primaryClass={cs.LG},
      url={https://arxiv.org/abs/2107.03374}, 
}

@article{wang2024mmlupro,
  title={Mmlu-pro: A more robust and challenging multi-task language understanding benchmark},
  author={Wang, Yubo and Ma, Xueguang and Zhang, Ge and Ni, Yuansheng and Chandra, Abhranil and Guo, Shiguang and Ren, Weiming and Arulraj, Aaran and He, Xuan and Jiang, Ziyan and others},
  journal={Advances in Neural Information Processing Systems},
  volume={37},
  pages={95266--95290},
  year={2024}
}

@article{rein2023gpqa,
  title={Gpqa: A graduate-level google-proof q\&a benchmark},
  author={Rein, David and Hou, Betty Li and Stickland, Asa Cooper and Petty, Jackson and Pang, Richard Yuanzhe and Dirani, Julien and Michael, Julian and Bowman, Samuel R},
  journal={arXiv preprint arXiv:2311.12022},
  year={2023}
}

@article{cobbe2021gsm8k,
  title={Training verifiers to solve math word problems},
  author={Cobbe, Karl and Kosaraju, Vineet and Bavarian, Mohammad and Chen, Mark and Jun, Heewoo and Kaiser, Lukasz and Plappert, Matthias and Tworek, Jerry and Hilton, Jacob and Nakano, Reiichiro and others},
  journal={arXiv preprint arXiv:2110.14168},
  year={2021}
}

@article{austin2021mbpp,
  title={Program synthesis with large language models},
  author={Austin, Jacob and Odena, Augustus and Nye, Maxwell and Bosma, Maarten and Michalewski, Henryk and Dohan, David and Jiang, Ellen and Cai, Carrie and Terry, Michael and Le, Quoc and others},
  journal={arXiv preprint arXiv:2108.07732},
  year={2021}
}

@article{qwen2025qwen3,
  title={Qwen3 technical report},
  author={Yang, An and Li, Anfeng and Yang, Baosong and Zhang, Beichen and Hui, Binyuan and Zheng, Bo and Yu, Bowen and Gao, Chang and Huang, Chengen and Lv, Chenxu and others},
  journal={arXiv preprint arXiv:2505.09388},
  year={2025}
}

@misc{qwen2024qwen2,
      title={Qwen2 Technical Report}, 
      author={An Yang and Baosong Yang and Binyuan Hui and Bo Zheng and Bowen Yu and Chang Zhou and Chengpeng Li and Chengyuan Li and Dayiheng Liu and Fei Huang and others},
      year={2024},
      eprint={2407.10671},
      archivePrefix={arXiv},
      primaryClass={cs.CL},
      url={https://arxiv.org/abs/2407.10671}, 
}

@misc{qwen2025qwen25,
      title={Qwen2.5 Technical Report}, 
      author={An Yang and Baosong Yang and Beichen Zhang and Binyuan Hui and Bo Zheng and Bowen Yu and Chengyuan Li and Dayiheng Liu and Fei Huang and Haoran Wei and others},
      year={2025},
      eprint={2412.15115},
      archivePrefix={arXiv},
      primaryClass={cs.CL},
      url={https://arxiv.org/abs/2412.15115}, 
}

@article{dubey2024llama3,
  title={The llama 3 herd of models},
  author={Grattafiori, Aaron and Dubey, Abhimanyu and Jauhri, Abhinav and Pandey, Abhinav and Kadian, Abhishek and Al-Dahle, Ahmad and Letman, Aiesha and Mathur, Akhil and Schelten, Alan and Vaughan, Alex and others},
  journal={arXiv preprint arXiv:2407.21783},
  year={2024}
}

@article{jiang2023mistral,
  title={Mistral 7B},
  author={Albert Q. Jiang and Alexandre Sablayrolles and Arthur Mensch and Chris Bamford and Devendra Singh Chaplot and Diego de las Casas and Florian Bressand and Gianna Lengyel and Guillaume Lample and Lucile Saulnier and Lélio Renard Lavaud and Marie-Anne Lachaux and Pierre Stock and Teven Le Scao and Thibaut Lavril and Thomas Wang and Timothée Lacroix and William El Sayed},
  journal={arXiv preprint arXiv:2310.06825},
  year={2023}
}

@article{gemma2024,
  title={Gemma 2: Improving open language models at a practical size},
  author={Team, Gemma and Riviere, Morgane and Pathak, Shreya and Sessa, Pier Giuseppe and Hardin, Cassidy and Bhupatiraju, Surya and Hussenot, L{\'e}onard and Mesnard, Thomas and Shahriari, Bobak and Ram{\'e}, Alexandre and others},
  journal={arXiv preprint arXiv:2408.00118},
  year={2024}
}

@inproceedings{hendrycks2017baseline,
  title={A Baseline for Detecting Misclassified and Out-of-Distribution Examples in Neural Networks},
  author={Hendrycks, Dan and Gimpel, Kevin},
  booktitle={International Conference on Learning Representations (ICLR)},
  year={2017}
}

@article{lin1991divergence,
  title={Divergence Measures Based on the Shannon Entropy},
  author={Lin, Jianhua},
  journal={IEEE Transactions on Information Theory},
  volume={37},
  number={1},
  pages={145--151},
  year={1991}
}

@article{kullback1951information,
  title={On Information and Sufficiency},
  author={Kullback, Solomon and Leibler, Richard A.},
  journal={The Annals of Mathematical Statistics},
  volume={22},
  number={1},
  pages={79--86},
  year={1951}
}

@inproceedings{scheffer2001active,
  title={Active Hidden Markov Models for Information Extraction},
  author={Scheffer, Tobias and Decomain, Christian and Wrobel, Stefan},
  booktitle={Advances in Intelligent Data Analysis (IDA)},
  pages={309--318},
  year={2001}
}

@article{williams1989learning,
  title={A Learning Algorithm for Continually Running Fully Recurrent Neural Networks},
  author={Williams, Ronald J. and Zipser, David},
  journal={Neural Computation},
  volume={1},
  number={2},
  pages={270--280},
  year={1989}
}
